\pdfoutput=1

\documentclass[11pt]{article}
\usepackage{enumitem}
\usepackage[final]{ACL2023}
\usepackage{float}
\usepackage{graphicx} 
\graphicspath{{pic/}}
\usepackage{times}
\usepackage{xcolor}
\usepackage{latexsym}
\usepackage{amsmath}
\usepackage{longtable}
\usepackage[normalem]{ulem}
\usepackage{booktabs} 
\usepackage[T1]{fontenc}

\newcommand{\blankspace}{\rule{15cm}{0.15mm}}

\usepackage[utf8]{inputenc}
\usepackage{adjustbox}
\usepackage{microtype}
\usepackage{multirow}
\usepackage{easyReview}
\usepackage{tcolorbox}
\usepackage{inconsolata}
\usepackage{pifont} 
\usepackage{adjustbox} 
\usepackage{booktabs} 
\usepackage{amsmath} 
\usepackage{pifont} 
\usepackage{adjustbox} 
\usepackage{booktabs} 
\usepackage{xcolor} 
\usepackage{xcolor, colortbl}
\usepackage{listings}
\usepackage{makecell}
\usepackage{multirow}
\usepackage{array}
\usepackage{caption}
\usepackage{booktabs} 

\definecolor{darkgreen}{rgb}{0.0, 0.5, 0.0}

\definecolor{softblue}{RGB}{40, 176 , 244} 
\definecolor{softyellow}{RGB}{255, 255, 150} 
\definecolor{lightgreen}{RGB}{145, 204, 117}

\newcommand{\cmark}{\textcolor{darkgreen}{\ding{51}}} 
\newcommand{\xmark}{\textcolor{red}{\ding{55}}} 

\newcommand{\hmark}{\textcolor{blue}{\ding{51}\rotatebox[origin=c]{-6.2}{\kern-0.7em\ding{55}}}}

\newcommand{\ours}{\textsc{Elaboration}}
\newcommand{\ourdata}{\textsc{Elaborationset}}

\definecolor{darkgreen}{rgb}{0.0, 0.5, 0.0}

\title{ELABORATION: A Comprehensive Benchmark on \\Human-LLM Competitive Programming}

\author{
Xinwei Yang$^{\spadesuit\heartsuit}$, \quad
Zhaofeng Liu$^{\diamondsuit}$, \quad
Chen Huang$^{\spadesuit\heartsuit}$, \quad
Jiashuai Zhang$^{\spadesuit}$, \quad \\
\textbf{Tong Zhang} $^{\spadesuit\heartsuit}$, \quad
\textbf{Yifan Zhang} $^{\triangle}$, \quad
\textbf{Wenqiang Lei}$^{\spadesuit\heartsuit}$\thanks{ \quad Corresponding author.} \quad
\\
${\spadesuit}$ Sichuan University \quad ${\diamondsuit}$ Tianjin University of Science and Technology \\ 
${\heartsuit}$ Engineering Research Center of Machine Learning and Industry Intelligence,\\Ministry of Education, China \quad ${\triangle}$ Vanderbilt University \\
xinwei\_yang@stu.scu.edu.cn  \quad \{scu.zhangtong, huangc.scu\}@gmail.com \\ wenqianglei@scu.edu.cn \\ 
}

\begin{document}
\maketitle

\begin{abstract}
While recent research increasingly emphasizes the value of human-LLM collaboration in competitive programming and proposes numerous empirical methods, a comprehensive understanding remains elusive due to the fragmented nature of existing studies and their use of diverse, application-specific human feedback. Thus, our work serves a three-fold purpose: First, we present the first taxonomy of human feedback consolidating the entire programming process, which promotes fine-grained evaluation. Second, we introduce \ourdata, a novel programming dataset specifically designed for human-LLM collaboration, meticulously annotated to enable large-scale simulated human feedback and facilitate cost-effective real human interaction studies. Third, we introduce \ours, a novel benchmark to facilitate a thorough assessment of human-LLM competitive programming. With \ours, we pinpoint strengthes and weaknesses of existing methods, thereby setting the foundation for future improvement. Our code and dataset are available at \url{https://github.com/SCUNLP/ELABORATION}.
\end{abstract}

\section{Introduction}

Competitive programming presents a formidable challenge, as it requires mastery of four key stages: 1) the precise understanding of complex problems, 2) the strategic planning of efficient solutions, 3) the generation of effective code within strict constraints, 4) and the rigorous debugging of their implementations \cite{cormen2022introduction, huang2023competition, dale2014programming}. 
To mitigate this challenge, there has been a growing interest in utilizing large language models (LLMs) for automatic competitive programming tasks \cite{nijkamp2022codegen, li2023starcoder, roziere2023code, guo2024deepseek, ridnik2024code, lozhkov2024starcoder, 10.1007/978-981-97-5669-8_11}, assisting individuals in CS education and technical interview preparation.
However, these models have not yet demonstrated remarkable performance for practical utility \cite{yan2023codescope, li2023taco, jain2024livecodebench}.

Driven by this concern, recent research has shifted from relying solely on LLMs to explore Human-LLM Competitive Programming, a human-in-the-loop paradigm that leverages multi-turn human feedback to enhance LLM efficacy \cite{shi2024can,chae2024coffee,zheng2024opencodeinterpreter}. 
However, existing research have been somewhat fragmented, with studies employing various, scattered and application-specific human feedback. This fragmentation hinders a comprehensive understanding of effective Human-LLM collaboration in competitive programming \cite{shi2024can}. For instance, \citet{mozannar2023simulating} and \citet{wang2024rocks} focus on suggesting solution strategies, while \citet{zheng2024opencodeinterpreter} concentrate on conversational error identification. These approaches overlook the potential advantages of human guidance in areas such as problem comprehension, solution planning. \textit{A comprehensive benchmark is therefore needed to evaluate the effectiveness and characteristics of human-LLM collaboration across the entire competitive programming process.}

\begin{figure*}[!h]
	\centering
	\includegraphics[width=0.99\textwidth]{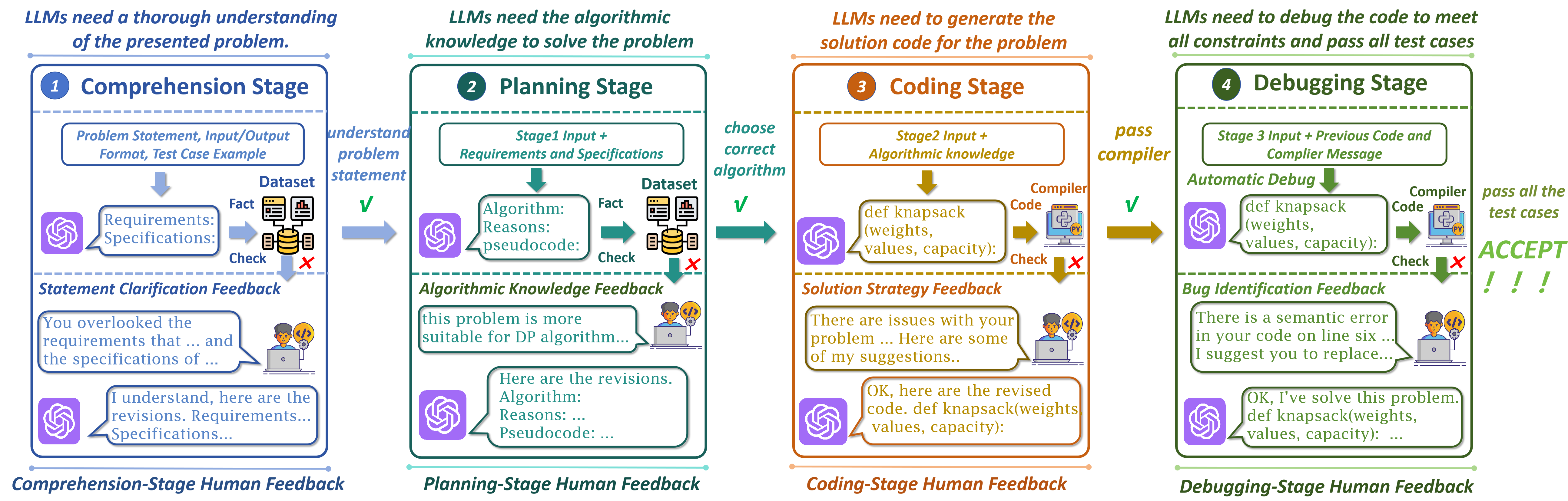}
    \setlength{\abovecaptionskip}{0pt}
    \setlength{\belowcaptionskip}{0pt}
	\caption{Illustration of \ours~evaluation. A human feedback taxonomy, structuring the entire programming process into four stages, enables stage-specific evaluation. }
    \vspace{-3mm}
	\label{fig:2}
\end{figure*}

To this end, we introduce \ours, a novel benchmark featuring a comprehensive evaluation protocol to facilitate a thorough assessment. This protocol incorporates a taxonomy of human feedback spanning the entire competitive programming process, and a new human-LLM programming dataset to support the evaluation implementations.
Specifically, our evaluation protocol builds upon existing works \cite{gao2024taxonomy,huang2024concept,chen2023travel}, using a conversational human-LLM interaction where textual human feedback is integrated into each code generation turn. As illustrated in Figure \ref{fig:2}, a novel taxonomy structures this human feedback, ensuring coverage across the entire competitive programming process: problem comprehension, solution planning, code generation, and debugging. 
This allows \ours~to incorporate human feedback at each stage and comprehensively assess its effectiveness. 
Moreover, to facilitate the evaluation implementation, we introduce \ourdata, the first competitive programming dataset specifically designed for human-LLM collaboration. This dataset comprises 8,320 problems from Codeforces and AtCoder, meticulously annotated to enable large-scale simulated human feedback and facilitate cost-effective real human interaction studies (cf. Table \ref{data_statistics}). As such, \ours~provides a robust and comprehensive framework for analyzing human-LLM competitive programming, paving the way for future advancements in this field.

\begin{table}[!ht]
\resizebox{\linewidth}{!}{
\centering
\begin{tabular}{lccc}
\toprule

\multicolumn{1}{l}{\textbf{Dataset}} & \textbf{Easy} & \textbf{Middle} & \textbf{Difficult} \\
\midrule
\multicolumn{4}{l}{\textbf{Basic Problem Information}} \\
Time Period & \multicolumn{3}{c}{Oct. 2011 \textasciitilde~Nov. 2024} \\
\#Problems &3642  &2098  &2580  \\
Avg. \#Test Cases &14.4  &14.5  &14.2 \\
\midrule
\multicolumn{4}{l}{\textbf{Annotations for Human Interaction (per Problem)}} \\
Avg. \#Statement Clarifications &8.1  &10.9  &12.1 \\
Avg. \#Algorithm Knowledge Summaries &2.4  &3.0  &3.8 \\
Avg. \#Ground Truth Solutions &4.8    &4.9    &4.8 \\ \midrule
\multicolumn{4}{l}{\textbf{Interaction Records with Real Humans}} \\
\#Problems  & 100 &100 &100 \\ 
Avg. \#Turns (\#Human Feedback) & 3.4 &5.1 &6.9    \\
Avg. \#Human-Annotated LLM Code Errors &1.3  &1.5  &2.0\\
\bottomrule

\end{tabular}}
\setlength{\abovecaptionskip}{1pt}
\setlength{\belowcaptionskip}{1pt}
\caption{\ourdata~Dataset statistics. Further details and examples are provided in Appendix \ref{dataset_exp}.}
\vspace{-3mm}
\label{data_statistics}
\end{table}

With \ours, we evaluate strengths and weaknesses of existing methods using both LLM-based user simulators and real human participants. Our findings demonstrate limited capacity of LLM alone for solving competitive programming problems, particularly those of high difficulty or unseen ones (-9.2\%, on average). Notably, human-LLM collaboration significantly improves task performance (+7.0\%, on average), particularly during the coding stage, although this requires efficient resource management. Real human experiments further highlight the complementary strengths of human and LLM bug identification, leading to a powerful synergy. In this paper, \ours~stands as a valuable resource to provide guidance and insight into benchmarking human-LLM competitive programming for future improvements. In conclusion, our contributions are as follows:
\begin{itemize}[label=\textbullet, leftmargin=*, itemindent=0.05cm, itemsep=-2pt]

    \item We introduce \ours, a novel benchmark for Human-LLM competitive programming, which features a comprehensive evaluation protocol to facilitate a thorough assessment.
    
    \item We present the first taxonomy of human feedback consolidating the entire programming process into four stages, enabling \ours~to evaluate task effectiveness at each stage.
    
    \item We introduce \ourdata, a novel programming dataset specifically designed for human-LLM collaboration. It comprises 8,320 problems, meticulously annotated to enable large-scale simulated human feedback and facilitate cost-effective real human interaction studies.

    \item With \ours, we evaluate pros and cons of existing methods using both LLM-based user simulators and real human participants, providing guidance and insight for future improvements.

\end{itemize}

\begin{table*}[ht]
\centering
    \begin{adjustbox}{width=1\textwidth}
    \begin{tabular}{l|c|ccc|ccc|cc}
    \toprule
\multirow{3}{*}{\textbf{\begin{tabular}[c]{@{}c@{}}Competitive Programming \\Benchmark\end{tabular} }} & \multirow{3}{*}{\textbf{\begin{tabular}[c]{@{}c@{}}Task Type\end{tabular}}} & \multicolumn{3}{c|}{\textbf{Basic Problem Information}} & \multicolumn{3}{c|}{\textbf{Annotations for Human Interaction}} & \multicolumn{2}{c}{\textbf{Real Human Interaction}} \\ \cline{3-10}
 & & \textbf{\begin{tabular}[c]{@{}c@{}}Contamination\\ Annotation\end{tabular}} & \multicolumn{1}{c}{\textbf{\begin{tabular}[c]{@{}c@{}}Stage\\ Annotation\end{tabular}}} & \multicolumn{1}{c|}{\textbf{\begin{tabular}[c]{@{}c@{}}Compile\\ Feedback\end{tabular}}} & \textbf{\begin{tabular}[c]{@{}c@{}}Clarify\\ Problem\end{tabular}} & \multicolumn{1}{c}{\textbf{\begin{tabular}[c]{@{}c@{}}Algorithmic\\ Knowledge\end{tabular}}} & \multicolumn{1}{c|}{\textbf{\begin{tabular}[c]{@{}c@{}}Ground Truth\\ Solutions\end{tabular}}} & \textbf{\begin{tabular}[c]{@{}c@{}}Bug\\ Annotation\end{tabular}} & \multicolumn{1}{c}{\textbf{\begin{tabular}[c]{@{}c@{}}Human-LLM\\ Multi-turn Records\end{tabular}}} \\

        
    \midrule
    APPS\cite{hendrycksapps2021} & Automatic & \xmark & \xmark & \xmark & \xmark & \xmark & \cmark & \xmark & \xmark\\
    CODE-CONTESTS\cite{li2022competition} & Automatic & \xmark & \xmark & \xmark & \xmark & \xmark & \cmark & \hmark & \xmark\\
    XCODEEVAL\cite{khan2023xcodeeval} & Automatic &  \xmark & \hmark & \xmark & \xmark & \xmark & \cmark & \hmark & \xmark\\
    CODESCOPE\cite{yan2023codescope} & Automatic &  \xmark & \hmark & \cmark & \cmark & \xmark & \hmark & \xmark & \xmark\\
    KareCoder\cite{huang2024knowledge} & Automatic&  \hmark & \xmark & \xmark & \xmark & \cmark & \cmark & \xmark & \xmark\\
    TACO\cite{li2023taco} & Automatic &  \xmark & \xmark & \xmark & \xmark & \xmark & \xmark & \xmark & \xmark \\
    USCAOBENCH\cite{shi2024can} & Automatic & \xmark & \hmark & \xmark & \xmark & \xmark & \cmark & \xmark & \xmark\\
    LIVECODEBENCH\cite{jain2024livecodebench} &Automatic & \cmark & \hmark & \xmark & \xmark & \xmark & \cmark & \xmark & \xmark\\
    \midrule
    OpenCoderInterpreter\cite{zheng2024opencodeinterpreter} &Human-LLM& \xmark & \xmark & \cmark & \xmark & \xmark & \hmark & \hmark & \xmark \\ 
    \textbf{\ours} (\textit{ours}) &Human-LLM & \cmark & \cmark & \cmark & \cmark & \cmark & \cmark & \cmark & \cmark\\
    \bottomrule
  \end{tabular}
  \end{adjustbox}
  \setlength{\abovecaptionskip}{1pt}
  \setlength{\belowcaptionskip}{1pt}
    \caption{Difference between \ours~and existing benchmarks. Only OpenCoderInterpreter and ours are specifically designed for human-LLM competitive programming. Here, '\hmark'~indicates partial support.}
      \label{table:ranking}
      \vspace{-3mm}
\end{table*}

\section{Related Work}
Our research focuses on human-LLM competitive programming, offering a comprehensive literature review and highlighting our novel contributions.

\noindent\textbf{Competitive Programming}. Competitive programming challenges participants to solve complex algorithmic problems under strict time and memory constraints \cite{dale2014programming}. Each problem begins with a detailed statement outlining the requirements and input/output specifications \cite{becker2023programming}. Unlike other programming tasks that focus on real-world applications, maintainability, readability, and collaboration \cite{passos2011turning, gallmeister1995posix, martin2003agile}, competitive programming demands precise problem comprehension, efficient algorithmic design, accurate code implementation, and thorough debugging to produce a solution that passes rigorous testing within the specified time and memory limits \cite{huang2023competition, dale2014programming, jain2024livecodebench}.

\noindent\textbf{Human-LLM Competitive Programming}. While the success of LLMs in other domains \cite{zhang2025enhancing, huang2025enableeffectivecooperationhumans} has fueled the application to automate competitive programming, recent benchmarks reveal limitations in their ability to solve expert-level problems \cite{hendrycks2021measuring, li2022competition, zheng2023survey, yan2023codescope, jain2024livecodebench}, even with compiler feedback (e.g., an error message) \cite{yang2024intercode, phung2023generating, tian2024debugbench}. This suggests limited practical utility when relying solely on LLMs for this complex task. Consequently, research is increasingly focusing on human-LLM competitive programming, which leverages multi-turn human feedback to enhance LLM performance. However, current methods often restrict human feedback to providing (pseudo-)code \cite{mozannar2023simulating,wang2024rocks} or debugging assistance \cite{zheng2024opencodeinterpreter,shi2024can}, neglecting the broader potential of human guidance across the entire programming process. This leads to fragmented understanding of the effectiveness and characteristics of human-LLM competitive programming, and further motivates our work.

\noindent\textbf{Human Feedback Simulation in Human-LLM Competitive Programming}. Evaluating any interactive systems is inherently labor-intensive \cite{huang-etal-2023-reduce}. Therefore, human simulators are commonly used in this field. While rule-based simulators have been employed \cite{mozannar2023simulating, zheng2024opencodeinterpreter}, their limited realism and comprehensiveness fall short of capturing the nuanced aspects of human feedback in competitive programming, which requires deep problem understanding, algorithmic knowledge, adaptive problem-solving, and error correction skills \cite{robins2003learning, pless2011introduction, lee2018understanding}. The emergence of LLM-based simulators offers a more realistic alternative, enhancing both simulation and evaluation reliability \cite{zheng2024opencodeinterpreter, mozannar2023simulating}. In line with these studies, our evaluation protocol also leverages LLM-based simulators to mimic human feedback at each programming stage. Crucially, we augment our benchmark with real human participants, providing a more grounded assessment.

\section{\ours\ Benchmark}



Our \ours \ benchmark evaluates human-LLM competitive programming using a novel protocol that incorporates a comprehensive taxonomy of human feedback, covering all stages of the process, and a new human-LLM programming dataset.

\noindent\textbf{Evaluation Protocol Overview}. 
Our evaluation protocol accommodates both real human programmers and user simulators to provide feedback at each stage of the competitive programming process, as illustrated in Figure \ref{fig:2}. For simplicity, we refer to both as "humans" unless otherwise noted. In this human-LLM competitive programming process, each LLM interacts iteratively with a human, generating intermediate results and receiving feedback until a correct solution is produced or a maximum number of iterations is reached. A correct solution is defined as code that passes all test cases within the specified time and memory limits. 

\subsection{Human Feedback Taxonomy}
To support comprehensive benchmark, we establish a taxonomy of human feedback, informed by analyses of real-world human interactions \cite{robins2003learning, fincher1999we} and established competitive programming practices \cite{cormen2022introduction, huang2023competition, dale2014programming}. This taxonomy consolidates the entire programming process into the following primary stages.
\begin{itemize}[leftmargin=*, itemindent=0.05cm, itemsep=-2pt]
    \item \textbf{Problem Comprehension}. LLMs require a thorough understanding of the problem statement. To facilitate this, human feedback can provide crucial requirements and specifications. For example, specifying edge cases that need to be handled (\textit{e.g., handling empty input arrays}), summarizing the functionalities that the code needs to implement (\textit{e.g., return the median value}), or highlighting the key constraints and objectives (\textit{e.g., solution must run in O(nlogn) time}).
    \item \textbf{Solution Planning}. LLMs engage in solution planning by selecting appropriate algorithms. To facilitate this, human feedback can suggest effective algorithms, provide justifications, and even supply complete and accurate pseudocode. For example, a human might suggest using Dijkstra algorithm for a shortest path problem, explaining its suitability for weighted graphs and providing the pseudocode for its implementation.
    \item \textbf{Code Generation}. LLMs must generate complete, compilable code. In this case, human feedback can suggest solution strategies to improve the generated code by, for example, suggesting a more efficient data structure (\textit{e.g., a stack}) and explicitly coding algorithm implementation details (\textit{e.g., using a binary heap-based priority queue and a stack for Dijkstra algorithm}).
    \item \textbf{Code Debugging}. LLMs must pass the complete set of test cases\footnote{It could use the compiler output to refine its code when necessary.}. In this case, humans assist in identifying errors until all unseen test cases are passed (\textit{e.g., pinpointing logic flaws leading to infinite loops}). Current, most exiting methods limit their focus at this stage and provide conversational feedback for error identification \cite{zheng2024opencodeinterpreter, shi2024can}.
\end{itemize}

\subsection{\ourdata~Dataset}
To facilitate our evaluation, we created \ourdata, a high-quality human-LLM programming dataset. It comprises 8,320 problems from Codeforces\footnote{\url{https://codeforces.com/}} and AtCoder\footnote{\url{https://atcoder.jp/}} between October 2011 and November 2024, meticulously annotated to enable both large-scale simulated human feedback and cost-effective studies using real human participants across all stages of the programming process. See Table \ref{data_statistics} and Figure \ref{fig:3} for illustration.

\noindent\textbf{Problem Information Collection}. Our dataset is collected in a three-step process: First, our automatic HTML scrapers\footnote{Scraper code will be released along with our dataset.} extract all necessary information from Codeforces and AtCoder, including problem statements, input/output formats, test case examples, dates, tags, and difficulty levels. Second, because not all code problems provide test cases, we utilize GPT-4o to generate them where needed, following the approach of \citet{li2023taco, jain2024livecodebench} and then manually check their validation. Third, the final dataset is split by date for our later contamination-free evaluation (i.e., evaluating the performance on unseen data).

\noindent\textbf{Annotations for Human Interaction}. To mitigate the labor-intensive and expertise-dependent nature of human problem-solving in competitive programming, \ourdata~incorporates fully accurate, static annotations for each problem. This provides a reliable reference for humans and facilitates cost-effective solutions for human-LLM collaboration. Specifically, annotations include: problem statement clarifications (requirements and specifications of each problem); algorithm-specific knowledge summaries (required algorithms to solve each problem and their definitions and pseudocodes); and ground truth solutions (see Figure \ref{fig:3}). This resource enables human reviewers to make informed feedback decisions and allows for the simulation of human participants with varying levels of expertise by adjusting the granularity of the provided feedback. Notably, all annotations, except ground truth solutions, undergo a two-stage process: initial LLM generation followed by manual review to ensure quality. Ground truth solutions are sourced directly from the respective programming platforms. Refer to Appendix \ref{sec:human_annotation} for details.

\section{Benchmark Experiments}
    



\subsection{Experiment Setup}
\noindent\textbf{Human Simulators}. Our benchmark incorporates LLM-based user simulators for large-scale evaluation, employing \textit{O1-Mini} to ensure realistic human simulation. In particular, we include the following two participant groups representing a range of programming expertise. By this means, we assess the effectiveness of the evaluated methods across a range of programming abilities and to understand how well the methods cater to different levels of user expertise. Notably, novice programmers are excluded due to their limited capacity to provide valuable feedback for LLM improvement. 
\begin{itemize}[leftmargin=*, itemindent=0.05cm, itemsep=-2pt]
    \item \underline{Student Programmer} (Intermediate Skill Level) possess more than basic programming knowledge but lack the deep expertise. Following established practices in human programmer simulation \cite{zheng2024opencodeinterpreter}, the \textit{O1-Mini} is prompted to provide feedback based on its internal knowledge.
    \item \underline{Teacher Programmer} (Expert Level) possess a high level of programming skill and experience. Unlike student programmer, this simulator leverages the complete \ourdata~dataset to ensure expert-level performance.
\end{itemize}

\noindent\textbf{Human Participants}. Our experiments also incorporate real human participants to gain practical insights. Refer to Section \ref{rq3} for details.

\noindent\textbf{LLM Models}. We benchmark thirteen LLMs, encompassing both closed-source and open-source models of varying sizes. This include O1-Mini \cite{o1}, GPT-4o \cite{gpt4o}, GPT-4-Turbo \cite{openai2023gpt}, Gemini-1.5-pro \cite{team2024gemini}, Claude-3.5 \cite{claude3.5sonnet_blog}, CodeLlama \cite{roziere2023code}, Deepseek-Coder \cite{guo2024deepseek}, Qwen2.5-Coder \cite{hui2024qwen2}.

\begin{figure}[t]
	\centering
	\includegraphics[width=0.47\textwidth]{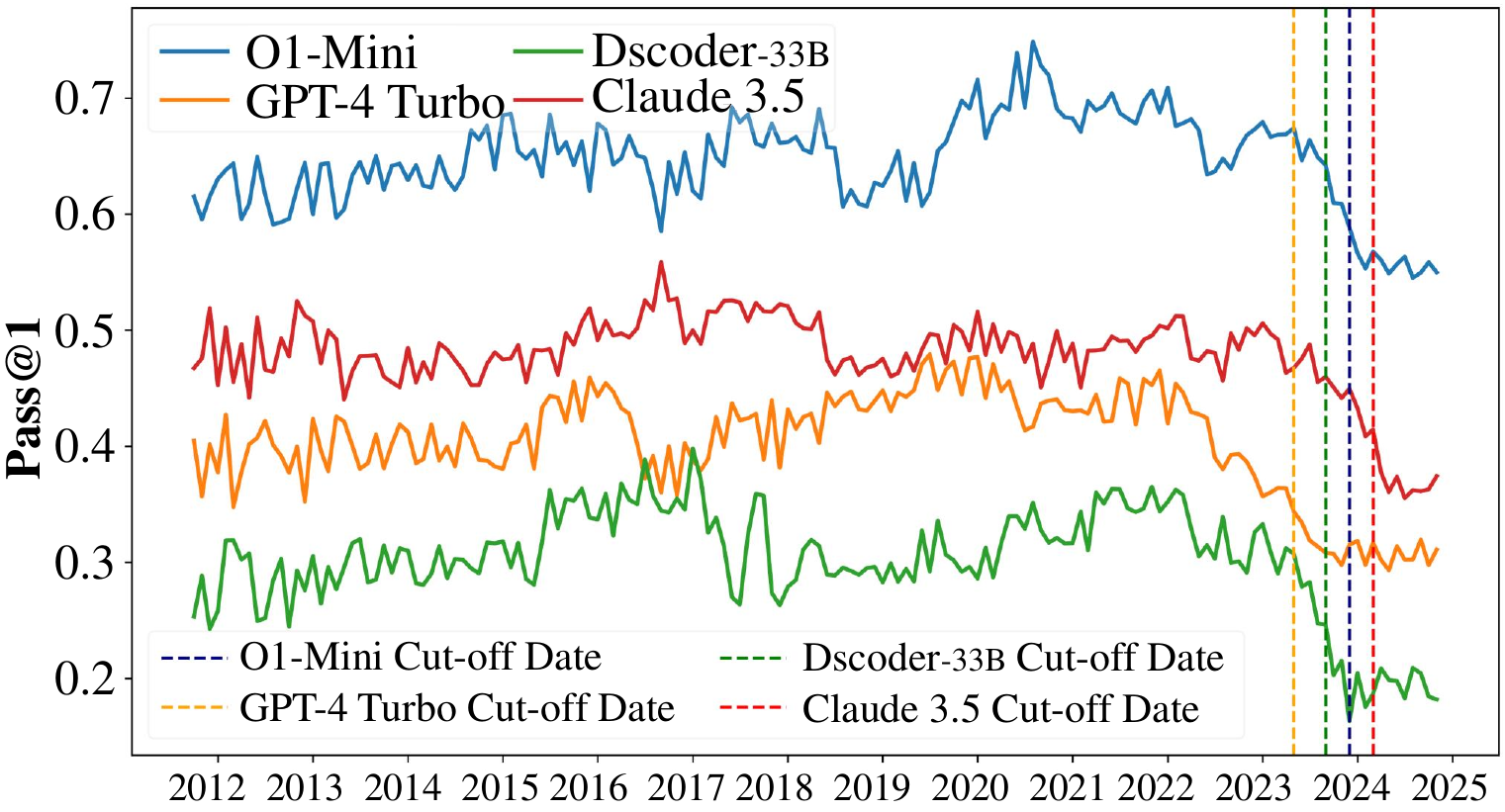}
    \setlength{\abovecaptionskip}{1pt}
     \setlength{\belowcaptionskip}{1pt}
	\caption{LLM Performance trends over time.}
	\label{fig:21}
    \vspace{-5mm}
\end{figure}

\noindent\textbf{Evaluation Metrics}. Following established practice \cite{humeval-2021-human, yan2023codescope, shi2024can, jain2024livecodebench}, we utilize the \underline{Pass@k} (k=1,3,5)\footnote{Given space limit, results with k=3,5 are in Appendix \ref{more}.} metric \cite{chen2021evaluating} to evaluate overall performance, with Pass@1 holding particular importance due to its relevance to real-world applications. 
To exclude the influence of potentially memorized solutions from the training corpus, we also employ a \uline{contamination-free evaluation}, focusing on problems released after the LLMs' respective cutoff dates.

\noindent\textbf{Implementation Details}. Our evaluation implementation proceeds through the forementioned four stages, with iterative human feedback provided until the human is satisfied with the LLM's response or a maximum iteration limit is reached. Fine-grained evaluation involves assessing LLM performance at each stage by comparing their outputs (e.g., summarized problem requirements and specifications, algorithm selection with justification, and pseudocode) against the annotated ground truth in our dataset. Code generation and debugging are evaluated based on final code performance, with error analysis conducted using either human participants or simulators. In our experiments, we utilize nucleus sampling, with a maximum of 10 iterations per stage. See Appendix \ref{detailsimp} for more details.


\begin{table*}[t]
\small
\centering
\resizebox{0.97\textwidth}{!}{
\begin{tabular}{@{}l|cccc|cccc@{}}
\toprule
\multirow{2}{*}{\textbf{Model} (Cut-off Date|Release Date)}
& \multicolumn{4}{c|}{\textbf{Contamination Evaluation} (\%)} &\multicolumn{4}{c}{\textbf{Contamination-free Evaluation} (\%)} \\ \cline{2-9}
& \textbf{Easy} & \textbf{Middle} & \textbf{Hard} & \textbf{Overall} & \textbf{Easy} & \textbf{Middle} & \textbf{Hard} & \textbf{Overall}  \\ \midrule

\textbf{O1-Mini} (2023-12 | 2024-09) & 88.1  & 70.3 & 41.7  & 66.7  & 80.6 & 66.6 & 30.8 & 59.3 \\
\midrule
\textbf{GPT-4o} (2023-11 | 2024-05) & 80.4 & 50.5 & 20.8 & 50.6 & 74.1 & 31.7 & 10.3 & 38.7   \\
\quad + Student Programmer Feedback  &83.1  &53.1  &24.3  &53.5  &76.2  &34.8  &15.1  &42.0   \\ 
\quad + Teacher Programmer Feedback  &87.7  &66.1  &38.2  &64.0  &80.1  &42.9  &23.3  &48.8 \\  \cline{2-9}
\textbf{GPT-4-Turbo} (2023-05 | 2023-11) &70.5  &40.6  &8.7  &39.9  &65.2 & 27.3 & 5.8  &32.8  \\

\quad + Student Programmer Feedback  &75.5  &46.1  &12.1  &44.6  &70.8  &33.2  &8.8  &37.6  \\ 
\quad + Teacher Programmer Feedback  &83.2  &58.8  &20.1  &54.0  &75.3  &39.8  &14.3  &43.1  \\  \cline{2-9}

\textbf{Gemini-1.5-pro} (2023-11 | 2024-02)  &81.2  &48.2  &22.0  &50.5  &73.2 &32.8 &9.3  &38.4  \\

\quad + Student Programmer Feedback  &84.0  &50.1  &25.1  &53.0  &75.5  &35.0  &13.1  &41.2  \\ 

\quad + Teacher Programmer Feedback  &89.1  &65.6  &36.6 &63.8  &81.0  &40.2  &24.2 & 48.5   \\  \cline{2-9}

\textbf{Claude-3.5} (2024-03 | 2024-06)  &78.0  &51.3  &16.2  &48.5  &74.5 &34.3 &5.4  &38.1  \\

\quad + Student Programmer Feedback  &82.2  &55.0  &24.1  &53.8  &76.6  &37.1  &7.9  &40.5  \\ 

\quad + Teacher Programmer Feedback  &87.0  &66.7  &33.4  &62.4  &83.1  &44.2  &16.5  &47.9   \\  \cline{2-9}

\textbf{\textit{Avg.}}  &77.5   &47.7   &16.9   &47.4   &71.8   &31.5   &7.7   &37.0  \\ 

\quad + Student Programmer Feedback  &81.2 (+3.7)   &51.1 (+3.4)  &21.4 (+4.5)   &51.2 (+3.8)  &74.8 (+3.0)   &35.0 (+3.5)  &11.2 (+3.5)  &40.3 (+3.3)  \\ 

\quad + Teacher Programmer Feedback  &86.8 (+9.3)  &64.3 (+16.6)  &32.1 (+15.2)  &61.1 (+13.7)  &79.9 (+8.1)  &41.8 (+10.3)  &19.6 (+11.9)  &47.1 (+10.1)    \\ \midrule

\multicolumn{9}{c}{\textbf{$\sim$7B Scale}} \\ \midrule

\textbf{CodeLlama-7B} (2023-01 | 2024-01)  & 30.3 & 5.9 & 0.5 & 12.2 & 15.2 & 2.1 & 0.3 & 5.9  \\

\quad + Student Programmer Feedback  &36.7  &10.3  &2.2  &16.4  &24.2  &3.1  &1.4  &9.6  \\ 

\quad + Teacher Programmer Feedback  &48.6  &17.8  &6.9  &24.4  &35.9  &8.4  &4.7   &16.3  \\  \cline{2-9}

\textbf{Deepseek-Coder-6.7B} (2023-09 | 2023-11)  &40.6  &15.4  &1.8  &19.3  &21.4  &7.0  &0.7  &9.7   \\

\quad + Student Programmer Feedback  &46.3  &18.8  &4.3  &23.1  &27.8  &11.3  &2.0  &13.7  \\ 

\quad + Teacher Programmer Feedback  &58.6  &27.8  &8.2  &31.5  &39.2  &24.2  &6.1  &23.2 \\  \cline{2-9}

\textbf{Qwen2.5-Coder-7B} (2024-06 | 2024-11)  &61.2  &22.4  &4.9  &29.5  &48.6  &9.3  &0.5  &19.5   \\

\quad + Student Programmer Feedback  &70.1  &26.6  &6.7  &34.5  &53.8  &12.3  &2.3  &22.8  \\ 

\quad + Teacher Programmer Feedback  &76.3  &35.5  &11.3  &41.0  &57.8  &21.6 &5.9  &28.4   \\  \cline{2-9}

\textbf{\textit{Avg.}}  &44.0  &14.6  &2.4  &20.3  &28.4  &6.1  &0.5  &11.7   \\

\quad + Student Programmer Feedback  &51.0(+7.0)  &18.6(+4.0)  &4.4(+2.0)  &24.7(+4.4)  &35.3(+6.9)  &8.9(+2.8)  &1.9(+1.4)  &15.4(+3.7)    \\ 

\quad + Teacher Programmer Feedback  &61.2(+17.2)  &27.0(+12.4)  &8.8(+6.4)  &32.3(+12.0)  &44.3(+15.9)  &18.1(+12.0)  &5.6(+5.1)  &22.6(+10.9)   \\ \midrule

\multicolumn{9}{c}{\textbf{$\sim$13B Scale}} \\ \midrule
\textbf{CodeLlama-13B} (2023-01 | 2024-01)  &35.8  &7.3  &1.7  &14.9  &23.5 &3.0  &0.3  &8.9   \\

\quad + Student Programmer Feedback  &40.3  &12.1  &2.9  &18.4  &26.3  &9.8  &1.4  &12.5  \\ 

\quad + Teacher Programmer Feedback  &44.2  &19.9  &5.8  &23.3  &29.8  &14.6  &3.1  &15.8   \\  \cline{2-9}

\textbf{Qwen2.5-Coder-14B} (2024-06 | 2024-11)  &70.8  &28.7  &7.7  &35.7  &58.3  &15.1  &2.2  &25.2   \\

\quad + Student Programmer Feedback  &75.9  &33.5  &10.2  &40.0    &61.2  &18.9  &4.1  &28.1  \\ 

\quad + Teacher Programmer Feedback  &80.1  &41.5  &14.2  &45.3  &66.3  &24.3  &6.8  &32.5   \\  \cline{2-9}

\textbf{\textit{Avg.}}  &53.3  &18.0  &4.7  &25.3  &40.9  &9.1  &1.3  &17.1   \\

\quad + Student Programmer Feedback  &58.1 (+4.8)  &22.8 (+4.8)  &6.6 (+1.9)  &29.2 (+3.9)  &43.8 (+2.9)  &14.4 (+5.3)  &2.8 (+1.5)  &20.3 (+3.2)    \\ 

\quad + Teacher Programmer Feedback  &62.2 (+8.9)  &30.7 (+12.7)  &10.0 (+5.3)  &34.3 (+9.0)  &48.1 (+7.2)  &19.5 (+10.4)  &5.0 (+3.7)  &24.2 (+7.1)   \\ \midrule

\multicolumn{9}{c}{\textbf{$\sim$34B Scale}} \\ \midrule

\textbf{CodeLlama-34B} (2023-01 | 2024-01)  &38.1  &7.9  &3.1  &16.4  &25.0  &5.1  &1.0  &10.4   \\

\quad + Student Programmer Feedback  &42.0  &12.3  &4.0  &19.4  &26.1  &8.4  &2.3  &12.3  \\ 

\quad + Teacher Programmer Feedback  &49.2  &18.8  &6.2  &24.7  &32.2  &13.0  &4.4  &16.5   \\  \cline{2-9}

\textbf{Deepseek-Coder-33B} (2023-09 | 2023-11)  &63.9  &23.7  &4.2  &30.6  &50.6  &10.4  &1.2  &20.7    \\

\quad + Student Programmer Feedback  &74.8  &28.7  &7.0  &36.8  &55.8  &13.3  &3.1  &24.0  \\ 

\quad + Teacher Programmer Feedback  &78.9  &40.1  &12.3  &43.8  &68.8  &20.4  &5.5  &31.6 \\  \cline{2-9}

\textbf{Qwen2.5-Coder-32B} (2024-06 | 2024-11)  &77.3  &41.3  &9.0  &42.5  &70.1  &20.3  &3.2  &31.2   \\

\quad + Student Programmer Feedback  &80.4  &45.3  &11.0    &45.6 &72.0  &23.1  &4.0 &33.0 \\ 

\quad + Teacher Programmer Feedback  &85.1  &53.4  &15.8  &51.4  &76.8  &30.1  &7.6  &38.0\\  \cline{2-9}

\textbf{\textit{Avg.}}  &59.8  &24.3  &5.4  &29.8  &48.6  &11.9  &1.8  &20.8   \\

\quad + Student Programmer Feedback  &65.7(+5.9)  &28.8(+4.5)   &7.3(+1.9)  &33.9(+4.1)  &51.3(+2.7)  &14.9(+3.0)  &3.1(+1.3)  &23.1(+2.3)    \\ 

\quad + Teacher Programmer Feedback  &71.1(+11.3)  &37.4(+13.1)  &11.4(+6.0)  &40.0(+10.2)  &59.3(+10.7)  &21.2(+9.3)  &5.8(+4.0)  &28.7(+7.9)   \\ \midrule

\textbf{\textit{Average over All LLMs}} & 60.7 & 28.6 & 8.4 & 32.6 & 50.0 & 16.5 & 3.4 & 23.3 
   \\

\quad + Student Programmer Feedback & 65.9(+5.2) & 32.7(+4.1) & 11.2(+2.8) & 36.6(+4.0) & 53.9(+3.9) & 20.0(+3.5) & 5.5(+2.1) & 26.4(+3.1) \\ 

\quad + Teacher Programmer Feedback & 72.3(+11.6) & 42.7(+14.1) & 17.4(+9.0) & 44.1(+11.5) & 60.5(+10.5) & 27.0(+10.5) & 10.2(+6.8) & 32.6(+9.3)
   \\

 \bottomrule
\end{tabular}} 
\setlength{\abovecaptionskip}{1pt}
\setlength{\belowcaptionskip}{1pt}
\caption{Pass@1 scores across various LLMs and varying levels of human feedback expertise. Since O1-Mini is expensive and recently released, experiments with it have been deferred. Refer to Appendix \ref{more} for more results.}
\label{tab:overall_results}
\vspace{-5mm}
\end{table*}

\subsection{Overall Performance (RQ1)}
This section benchmarks the performance of human-LLM competitive programming, assessing both overall performance and performance within specific problem categories.
We report the results in Table \ref{tab:overall_results} and draw the following observations.

\noindent\textbf{\textit{Are LLMs qualified competitive programmers?} -- They demonstrate limited capacity for solving competitive programming problems, particularly those of high difficulty or unseen ones}. As shown in Table \ref{tab:overall_results}, model performance exhibits a positive correlation with parameter size (larger models generally perform better), with the recently released O1-Mini achieving the best results, with a pass@1 score of 59.3\% on unseen problems. However, this effectiveness is limited to simpler programming problems. Performance across all LLMs, including those specifically designed for coding tasks, degrades significantly as problem difficulty increases, with the average pass@1 score is merely 3.4\% on unseen hard problems, rendering them alone unsuitable for real-world applications. Furthermore, performance deteriorates even further in contamination-free evaluations, as illustrated in Table \ref{tab:overall_results} and Figure \ref{fig:21}, with an average drop of 9.3\% on unseen problems compared to seen ones. This suggests that a substantial portion of LLM performance may stem from memorization of the training dataset\footnote{This memorization isn't simply rote learning; LLMs still produce some correct answers on unseen problems.}, a issue warrants further investigation.

\noindent\textbf{\textit{How effective can human feedback be in assisting LLMs with competitive programming challenges?} -- Human-LLM collaboration significantly enhances LLM performance, demonstrating the crucial role of human feedback}. As shown in Table \ref{tab:overall_results}, the integration of human participation throughout the programming process, creating a human-LLM competitive programming framework, resulted in significant performance gains across various LLMs, problem difficulties, and levels of human expertise. Interestingly, such performance gain is consistently observed regardless of data contamination, definitively demonstrating the power of human-LLM collaboration in solving complex programming challenges. This human-LLM collaboration approach resulted in an average increase of 9.3\% in the Pass@1 score when teacher programmers provided feedback on unseen problems, and an average increase of 11.5\% when they offered feedback on seen problems. Similarly, student programmers contributed to an average improvement of 3.1\% in the Pass@1 score for unseen problems and 4.0\% for seen problems. However, the integration of human feedback necessitates a corresponding investment of human effort, a topic explored further in the following section. 


\begin{figure}[t]
	\centering
	\includegraphics[width=0.47\textwidth]{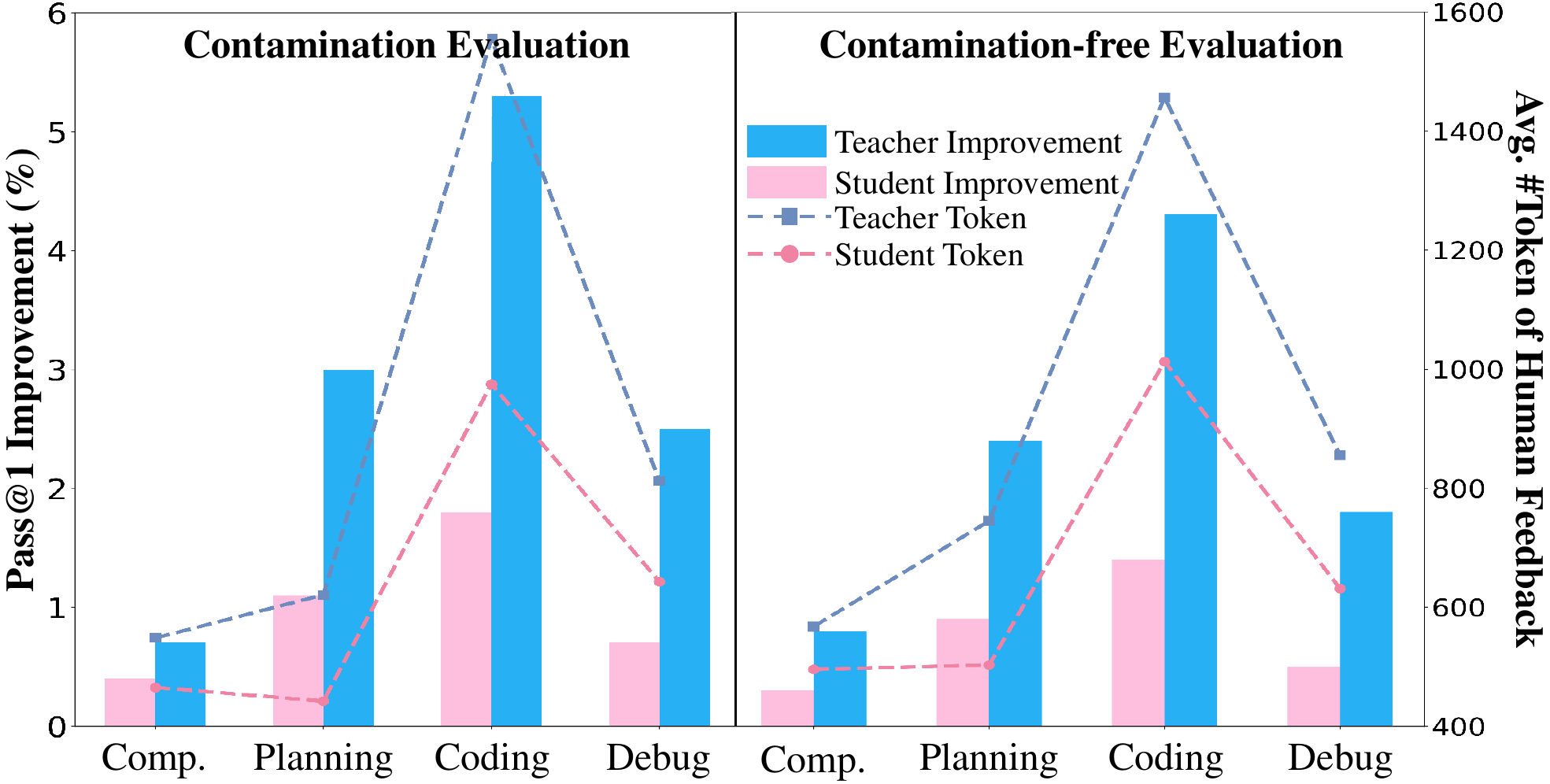}
    \setlength{\abovecaptionskip}{1pt}
\setlength{\belowcaptionskip}{1pt}
	\caption{Stage-specific evaluation averaged over various LLMs. While coding-stage feedback is most beneficial, it also incurs higher token usage.}
	\label{fig:19}
    \vspace{-3mm}
\end{figure}

\subsection{Finer-grained Analysis (RQ2)}
This section delves into the detailed characteristics of human-LLM competitive programming, with specical focus on the task performance and cost efficiency across various stages.

\noindent\textbf{\textit{At what stage of the programming process is human feedback most beneficial?} -- During the coding stage, even on problems with no data contamination}. Figures \ref{fig:19} illustrate the effectiveness of human feedback at different stages of competitive programming, both with and without data contamination, as measured by the average improvement in Pass@1. 
Regardless of data contamination, The results indicate that human feedback is consistently least effective during the comprehension stage and most effective during the coding stage, indicating that LLMs readily understand problem statements (cf. Table \ref{comprehension_planning}, high performance at comprehension stage) but struggle to generate correct code. Taking Table \ref{tab:feedback} for example, when tackling the classic 8-queens problem, the LLM frequently makes initialization errors and omits checks for queen conflicts. In this case, targeted human feedback during the coding stage can effectively mitigate these issues. Crucially, the minimal improvement observed with debugging-stage feedback highlights the importance of providing guidance throughout the entire process, underscoring our contributions. 

\begin{table}[t]
\centering
\resizebox{0.4\textwidth}{!}{
\begin{tabular}{lccc}
\toprule
\multicolumn{1}{l}{\textbf{Stage}} & \textbf{Easy} & \textbf{Middle} & \textbf{Hard} \\
\midrule
Comprehension Stage  &0.96  &0.93   &0.90 \\
Planning Stage &0.72  &0.53  &0.41  \\
\bottomrule
\end{tabular}}
\setlength{\abovecaptionskip}{1pt}
\setlength{\belowcaptionskip}{1pt}
\caption{Fine-grained evalution at comprehension and planning stages. We report averaged comprehension accuracy of summarized requirements and specifications, and average planning accuracy of algorithm selection. Refer to Appendix \ref{nuance} for nuanced understanding.}
\label{comprehension_planning}
\vspace{-5mm}
\end{table}

\begin{table}[h]
    \centering
    \resizebox{0.44\textwidth}{!}{%
    \begin{tabular}{p{0.97\linewidth}} 
        \toprule
        \textbf{\small{Teacher Programmer Feedback}} \\
        \midrule
        \small {To implement the 8-queens problem, start by initializing the board representation, usually as a one-dimensional array of length 8, with initial values set to a placeholder (like -1). Prepare auxiliary functions to verify the legality of queen placements and be ready to store potential solutions. When placing a queen in each row i from 0 to row -1, return False if there is a conflict with any previously placed queen in the same column or on either diagonal (the main diagonal from top-left to bottom-right or the secondary diagonal from top-right to bottom-left). Then, ensure that you assign the corresponding value in the array to the column number. Finally, if the row number equals the number of queens, return the array.}\\
        \midrule
        \textbf{\small{Student Programmer Feedback}} \\
        \midrule
        \small {When implementing the 8-queens problem, initialize the board representation, typically as a one-dimensional array of length 8, set initial values to a placeholder, prepare auxiliary functions to check the legality of queen placements, and be ready to store potential solutions. }\\
        \bottomrule
    \end{tabular}
    }
    \setlength{\abovecaptionskip}{1pt}
\setlength{\belowcaptionskip}{1pt}
    \caption{Coding-stage feedback comparison on 8-queens problem. Teacher feedback is more detailed with specific placeholder value, iterative placement strategy, and explicit backtracking, etc.}
    \label{tab:feedback}
    \vspace{-5mm}
\end{table}


\noindent\textbf{\textit{What are characteristics of different types of programmer feedback?} -- While detailed, expert feedback yields greater benefits, its higher cost necessitates efficient use of human resources}. As illustrated in Figures \ref{fig:19}, teacher programmers generally achieve higher Pass@1 improvement than student programmers, attributable to the more detailed and nuanced nature of their feedback. However, this improvement comes at a significant cognitive cost. For example, given the classic 8-queens problem, the student programmer feedback might miss several crucial details compared to teacher feedback (cf. Table \ref{tab:feedback}), such as specific placeholder value, iterative placement strategy, and explicit backtracking. 
Following previous studies \cite{owoicho2023exploiting,wu2024need}, we further calculate the average number of tokens in human feedback, identifying a substantial token overhead (indicated by the dashed line in the figures), particularly during the coding stage. While human participants in collaborative programming may be willing to invest time, the high cost necessitates more efficient methods for LLM integration of human feedback. Currently, a preliminary cost-benefit analysis (by Pass@1/\#token) suggests that planning-stage feedback might be more cost-effective than currently implemented. Therefore, future research within the community should prioritize the development of cost-effective methods for integrating human feedback to address this challenge.


\subsection{Collaborating with Real Humans  (RQ3)}
\label{rq3}
With \ourdata, we benchmark existing methods using real human programmers to gain practical insights into their characteristics.

\begin{table}[t]
\resizebox{\linewidth}{!}{
\renewcommand{\arraystretch}{1.4} 
\begin{tabular}{c|c|cc|cc}
\toprule
\multirow{2}{*}{\large \begin{tabular}[c]{@{}c@{}}Debug\\ Type\end{tabular}} & \multirow{2}{*}{\large {\begin{tabular}[c]{@{}c@{}}Difficulty\\ Level\end{tabular}}} & \multicolumn{2}{c|}{\large Error Identification} & \multicolumn{2}{c}{\large Problem Resolution (P@1)} \\ \cline{3-6}
& & \large Precision & \large Recall & \large Original & \large + Debug \\ \midrule
\multirow{4}{*}{  \begin{tabular}[c]{@{}c@{}}Automatic\\ Debug\end{tabular}  } 
& Easy    &0.34  &0.56  &  0.66 & 0.73 \\
& Middle    &0.22  & 0.36  &  0.27 & 0.33   \\
& Hard    &0.14  &0.28    &  0.06 & 0.08 \\  \cline{2-6}
& Overall  &0.23  &0.40    &0.33   &0.38  \\
\midrule

\multirow{4}{*}{\begin{tabular}[c]{@{}c@{}}Human\\ Debug\end{tabular}}   
& Easy  &0.92   &0.78    &0.73   &0.92  \\
& Middle  &0.80    &0.72  &0.33  &0.65       \\
& Hard  &0.72    & 0.64   &0.08 &0.29    \\ \cline{2-6}
& Overall  &0.81   &0.71    &0.38   &0.62  \\

\bottomrule
\end{tabular}}
\setlength{\abovecaptionskip}{1pt}
\setlength{\belowcaptionskip}{1pt}
\caption{Analysis of GPT-4 Turbo error identification and resolution with automatic and human debugging.}
\vspace{-5mm}
\label{Debug_score}
\end{table}

\noindent\textbf{Setup}. Five computer science graduate students are employed in this study. Following \citet{shi2024can,tian2024debugbench}, they only provide textual feedback identifying syntactic and semantic errors (cf. Table \ref{Bug_Type}) rather than direct code editing. The collaborative process continues until a correct solution is found or a maximum of 10 iterations is reached. Considering human labor, we focus humans on the debugging stage\footnote{Refer to Appendix \ref{codinghuman} for results on coding state.} using a subset of 300 randomly selected unseen problems from \ourdata. For the LLM, we consider GPT-4-Turbo (due to its balance of strong performance and reasonable cost). we allows GPT-4-Turbo to refine its solution based on both compiler feedback and simulator feedback played by O1-Mini (We term this process as \textbf{Automatic Debug}), which reduces the human workload for bug identification. Finally, \uline{we conduct a post-experiment review, where bugs within all generated codes are meticulously annotated. This supplementary dataset will be made publicly available along with our dataset}. Refer to Appendix \ref{setforhuamn} for details.

\noindent\textbf{\textit{How valuable is human-LLM collaboration from a practical perspective?} -- Humans play a vital role in identifying bugs and improving LLM performance}. Table \ref{Debug_score} reveals that automatic debug struggles to accurately identify bugs, achieving only 23\% precision and 40\% recall, resulting in a mere 5\% improvement in Pass@1 performance. In contrast, incorporating human bug identification significantly improved results, yielding 81\% precision and 71\% recall, and a substantial 24\% increase in Pass@1 performance, demonstrating the critical role of human intervention.


\noindent\textbf{\textit{How do human and LLM bug detection differ?} -- They have complementary strengths, creating a powerful synergy}. We conduct in-depth debug analysis and report the results in Table \ref{Bug_Type}. Our experiments show that GPT-4 Turbo generates significantly more semantic bugs than syntactic ones, especially incomplete and logically flawed errors. While automatic debugging effectively addresses most syntactic errors (nearly all when combined with human debugging), it struggles with semantic errors. Human debugging significantly improves the resolution of these semantic errors, particularly those involving references, calculations, incompleteness, and logical flaws. This highlights the complementary strengths of humans and LLMs, each identifying different types of errors \cite{rosenfeld2018totally}, and underscores the human-LLM collaboration for more accurate outputs.


\begin{table}[t]
\resizebox{\linewidth}{!}{
\begin{tabular}{c|c|ccc}
\toprule
\begin{tabular}[c]{@{}c@{}}Bug\\ Category\end{tabular} &Bug Type & Original & \begin{tabular}[c]{@{}c@{}}+ Automatic\\ Debug\end{tabular}  & \begin{tabular}[c]{@{}c@{}}+ Human\\ Debug\end{tabular} \\ \midrule
\multirow{5}{*}{ \begin{tabular}[c]{@{}c@{}}Syntactic\\ Bugs\end{tabular}  } 
&Function Related Errors     &11   &4 & 1    \\
&Operation Errors     &3   &1 &0       \\
&Structure Errors     &4   &1  &0        \\  
&Declaration Errors    &8   &2   &0      \\
&Import Errors     &7   &2  &0        \\  \cline{2-5}
&Overall  &33   &10   &1 \\
\midrule

\multirow{6}{*}{\begin{tabular}[c]{@{}c@{}}Semantic\\ Bugs\end{tabular}}   
&Control Flow Errors   &58  &46 &30      \\
&Reference Errors   &17  &16 & 3          \\
&Calculation Errors   &23  &25 &4          \\
&Incomplete Errors   &142  &99 & 55         \\
&Logical Direction Error   &87  &33 &12         \\
&Suboptimal Errors   &23  &20 &19         \\\cline{2-5}
&Overall &350  &239 &123 \\

\bottomrule
\end{tabular}}
\setlength{\abovecaptionskip}{1pt}
\setlength{\belowcaptionskip}{1pt}
\caption{Bug statistics for GPT-4 Turbo: with and without feedback. Bug description are in Table \ref{error_taxonomy}.}
\vspace{-5mm}
\label{Bug_Type}
\end{table}

\begin{figure}[h]
	\centering
	\includegraphics[width=0.3\textwidth]{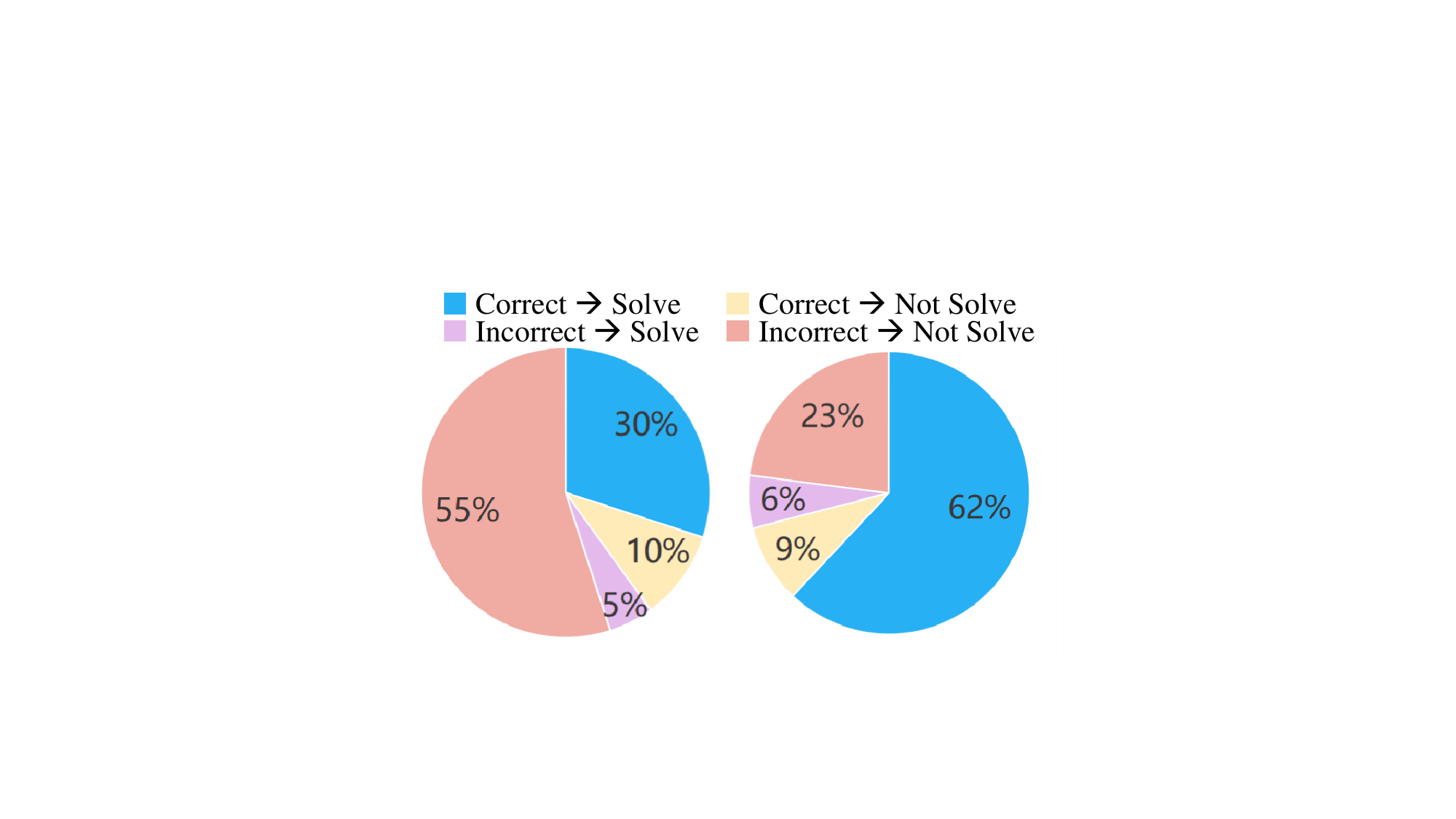}
    \setlength{\abovecaptionskip}{1pt}
\setlength{\belowcaptionskip}{1pt}
	\caption{Bug correction success rates via correct and incorrect automatic (\textit{left}) and human feedback (\textit{right}) .}
	\label{fig:10}
    \vspace{-5mm}
\end{figure}

\noindent\textbf{\textit{How effectively does LLM utilize different types of feedback?} -- It demonstrates higher success rates correcting bugs with accurate human feedback}. We analyze GPT-4 Turbo's effectiveness in utilizing automatic and human feedback for bug correction. As illustrated in Figure \ref{fig:10}, while LLMs struggle with initial bug identification, they demonstrate a strong capacity for correction when provided with accurate bug information. With accurate automatic bug identification, LLMs successfully resolve 75\% of bugs. This increases to 87\% with accurate human feedback. 
Based on our analysis, when GPT-4 Turbo failed to correct errors despite receiving accurate feedback, the errors are predominantly of Control Flow Errors and Suboptimal Errors. This indicates that direct human code modification may be necessary to resolve these error types.

\section{Conclustion}
We study the effectiveness and characteristics of human-LLM competitive programming by (1) introducing a novel taxonomy of human feedback for fine-grained evaluation; (2) providing \ourdata, a new dataset for human-LLM collaboration; and (3) developing \ours, a benchmark for evaluating off-the-shelf methods and identifying their pros and cons. Thus, our work stands out as a valuable resource to provide guidance for future improvement in this field.


\section*{Limitations}
\textbf{Sensitive to Prompts}. As with other LLM prompting studies\cite{zhang2024clamber,huang2024co,zhang2024strength,chen2024style}, our results may be sensitive to prompt. While our prompts underwent rigorous review and testing, and our main experiments report averages across over 8,000 problems, optimizing prompts for this specific task remains a significant challenge and area for future research.

\noindent\textbf{Generalizability to Other Programming Tasks}. In accordance with scientific rigor, this study defines its scope as Human-LLM collaboration within competitive programming, a domain chosen to examine the capabilities and limitations of both LLMs and human performance. While acknowledging the potential relevance to broader programming tasks, we limit our evaluations and analyses to this specific context and defer extending the representativeness of our results to general software development or other programming domains. Despite this focus, elements of our work offer valuable insights applicable to diverse programming scenarios. The problem-solving process shares fundamental similarities across programming contexts, and our proposed human feedback taxonomy and methods for improving problem comprehension in LLMs may readily translate. Developers, for example, can leverage clear and detailed feedback on specifications, as demonstrated in our benchmark, to guide LLMs towards a better understanding of software requirements. We believe this highlights pathways for broader applicability and welcome further discussion.

\section*{Ethics Statement} 
The proposed dataset for this study is primarily sourced from publicly available, reputable competitive programming websites. Our data collection process strictly avoids any personally identifiable information, such as user IDs, avatars, or comments, ensuring maximum transparency and accessibility. Furthermore, in our work, the dataset is manually annotated, and human-LLM collaborative programming is employed. During our experiments, we provide human participants with a full explanation of data usage and publication; at no point are participants exposed to inappropriate content. We ensure that the whole process adhered to all ethical guidelines and ensured the responsible and transparent use of human participants' time and effort, thereby promoting the advancement of research in the field. For all human-subject studies, we strictly adhered to IRB approval. To control the workload of the human annotators, we used a two-stage annotation process, starting with three LLMs(O1-mini) performing automatic annotation, followed by human annotators to control the quality. Specifically, human annotators would further annotate issues where there was disagreement in the automatic annotation by the LLMs. Two human annotators individually annotated 244 and 318 items, respectively. Participants in the human-subject study were compensated \$150 for their involvement, while human annotators were each compensated \$300 for their work.

\section*{Acknowledgements}
This work was supported in part by the National Natural Science Foundation of China (No. 62272330 and No.U24A20328); in part by the Fundamental Research Funds for the Central Universities (No. YJ202219); in part by the Science Fund for Creative Research Groups of Sichuan Province Natural Science Foundation (No. 2024NSFTD0035); in part by the National Major Scientific Instruments and Equipments Development Project of Natural Science Foundation of China under Grant (No. 62427820); in part by the Natural Science Foundation of Sichuan (No. 2024YFHZ0233)

\bibliography{anthology,custom}
\bibliographystyle{acl_natbib}

\newtcolorbox{promptbox}[2][Prompt]{
    colback=black!5!white,
    arc=5pt, 
    boxrule=0.5pt,
    fonttitle=\bfseries,
    title=#1, 
    before upper={\small}, 
    fontupper=\rmfamily, 
    colframe=#2, 
    width=\textwidth,   
    enlarge left by=0mm, 
    enlarge right by=0mm, 
}


\appendix

\section{Details of Dataset Description and Construction}
\label{dataset_exp}

\subsection{Dataset Description}
Our dataset include the following threefold information. This dataset will be openly released soon.

\begin{figure*}[ht]
	\centering
	\includegraphics[width=0.95\textwidth]{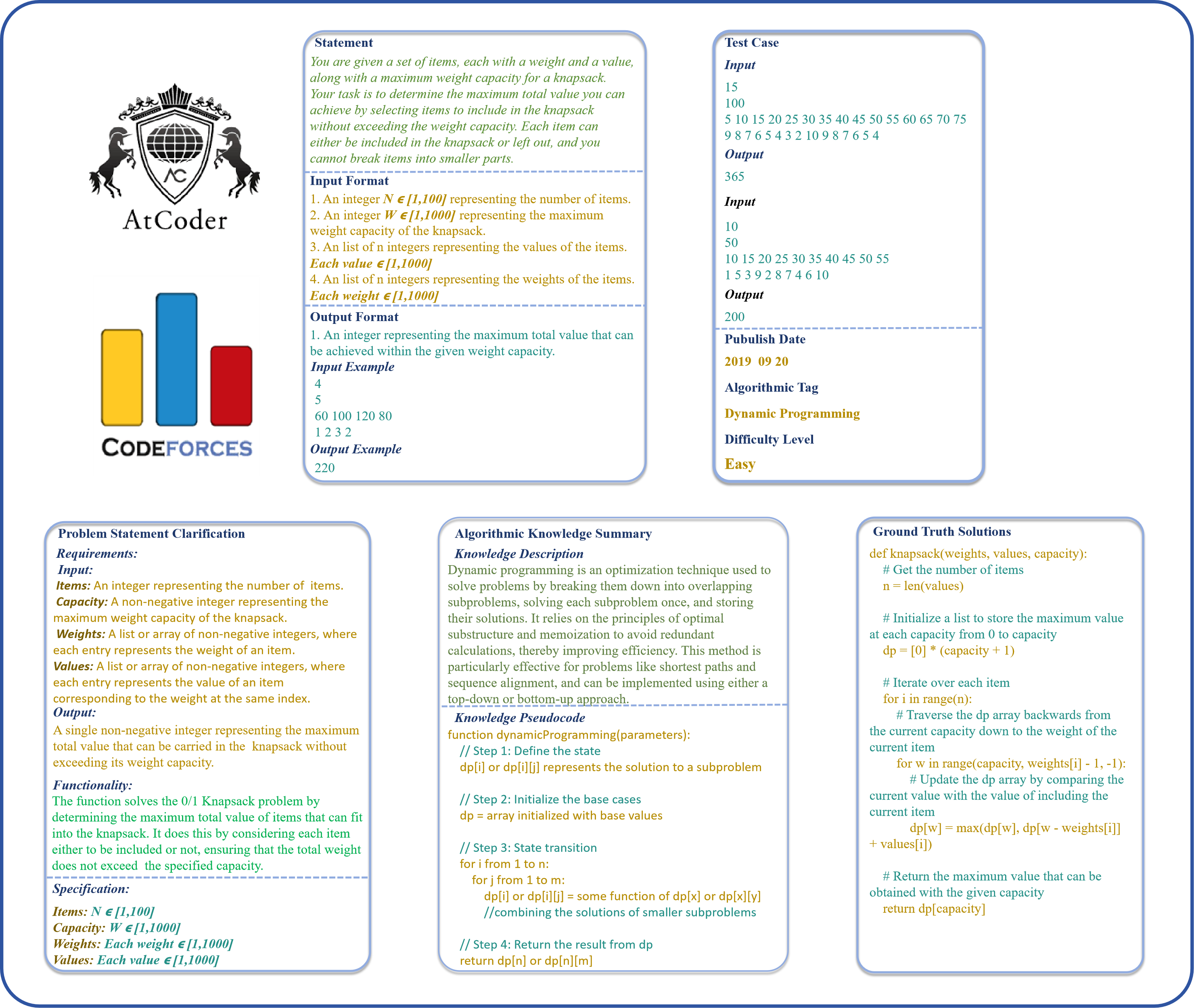}
	\caption{Description of dataset}
	\label{fig:3}
\end{figure*}

\noindent\textbf{Static Dataset}. As shown in Figure \ref{fig:3}, \ourdata \ is primarily composed of two parts: Problem Information Collection and Annotations for Human Interaction. The former includes problem statements, input/output formats, test cases, examples, dates, tags, and difficulty levels. The latter consists of carefully annotated problem statement clarifications, as well as algorithm-specific knowledge summaries, which include the required algorithms to solve each problem along with their definitions and pseudocodes.

\noindent\textbf{Human Simulator-LLM Competitive Programming Dataset}. Our dataset also includes multi-turn interaction data between 13 LLMs and two human simulators (emulated by O1-Mini), encompassing multi-turn feedback from human simualtor and LLM-generated codes. This data facilitates future research into LLM behavior.

\noindent\textbf{Real Human-LLM Competitive Programming Dataset}. Additionally, we also include multi-turn interaction data between GPT-4 Turbo and five real humans. This covers 300 problems of varying difficulty. See Appendix \ref{setforhuamn} for construction details.

\begin{table}[h]
\resizebox{\linewidth}{!}{
\centering
\begin{tabular}{@{}llcc@{}}
\toprule
\textbf{Dataset} & \textbf{Difficulty} & \textbf{Difficulty Level} & \textbf{Problem Number} \\ \midrule
Codeforces & Easy &(0,750]  & 2332 \\
 & Middle &(750,1000]  & 907 \\
 & Hard  & (1000,1500] & 1423 \\ \midrule
Atcoder & Easy &(0,350] &1310 \\
 & Middle &(350,550] &  1191\\
 & Hard  &(550,900] & 1157\\ \bottomrule
\end{tabular}}
\caption{\ourdata: Difficulty Level}
\label{difficulty_level}
\end{table}

\subsection{Static Dataset Construction}

\subsubsection{Problem Information Collection}

\noindent\textbf{Basic Problem Collection}. Following \citet{jain2024livecodebench}, we collected problem statements, input/output formats, example test cases, publication dates, algorithmic tags, and difficulty levels from publicly accessible sections of the AtCoder and Codeforces websites, removing any duplicates. Notably, we focus on scraping only the publicly accessible sections of these sites, steering clear of any data that may be behind paywalls or require user login or interaction. We will release our code along with our dataset.

\noindent\textbf{{Problem Difficulty Level}}. Codeforces and AtCoder assign difficulty scores to problems using points-based systems, with higher scores indicating greater difficulty. Codeforces categorizes problems as Easy, Medium, and Hard based on score ranges of (0, 750], (750, 1000], and (1000, 1500], respectively (Table \ref{difficulty_level}). AtCoder uses ranges of (0, 350], (350, 550], and (550, 900]. We excluded the most hard problems, as these are currently beyond the capabilities of LLMs.

\noindent\textbf{{Generaing Test Case When Necessary}}. While we collected test cases from both websites, we found that some problems lacked them, specially for problems from Codeforces. In response, we used GPT-4o to generate them, following the approach of \citet{li2023taco, jain2024livecodebench}. Using prompt \ref{Random_input}, we generated 15 diverse test case inputs per problem based on the corresponding problem statements, and then generated corresponding outputs using ground truth Python code. Three separate ground truth codes and human programmers validated the accuracy of these test cases.


\noindent\textbf{Fair Use and Academic Purpose.} In accordance with \citet{hendrycks2021measuring,jain2024livecodebench}, we adhere to Fair Use §107, which states that “the fair use of a copyrighted work, including such use by ... scholarship or research, is not an infringement of copyright.” Fair use is assessed based on “the purpose and character of the use, including whether it is of a commercial nature or for nonprofit educational purposes,” “the amount and substantiality of the portion used in relation to the copyrighted work as a whole,” and “the effect of the use upon the potential market for or value of the copyrighted work.” We emphasize that the problems we have collected are used solely for academic purposes, and we do not train on these collected problems.

\subsubsection{Annotations for Human Interaction}
\label{sec:human_annotation}

The annotation process aims to guarantee the precision and clarity of problem statements, algorithm-specific knowledge, and ground truth solutions. To accomplish this, we employ a multi-step approach. Initially, O1-mini carries out automatic annotation, followed by manual verification.

\noindent\textbf{Problem Statement Clarifications}. Initially, we utilize O1-Mini following the structured prompt \ref{tab:comprehension_prompt} to generate clarifications of problem statements for each issue. This includes refining ambiguous descriptions and outlining clear and concise requirements and specifications to guide problem-solving. We used three LLMs O1-Mini for annotation and conducted discussions. After the automatic generation step, we involved two master's students in computer science to perform a thorough manual evaluation of the generated content for those problems where there was no agreement in the results. These annotators verify the alignment of the requirements and specifications with the problem objectives and ensure that all critical aspects are covered. Their expertise helps ensure the annotations are precise, comprehensive, and aligned with academic standards.

\noindent\textbf{Knowledge Summaries Annotated.} For the algorithm-specific knowledge summaries, we compile detailed descriptions and pseudocode for 33 distinct algorithms. Initially, three LLMs O1-Mini are employed to generate drafts of these summaries, including algorithm definitions, purposes, and step-by-step pseudocode. Subsequently, for the results where the LLMs could not reach consensus, the generated outputs undergo meticulous manual review to verify the correctness of both the descriptions and the pseudocode. This verification process includes cross-referencing with standard algorithmic literature to ensure consistency and accuracy. Examples of these algorithm summaries are presented in Table \ref{knowledge_summary_examples}, showcasing both the variety and depth of the annotations.

\noindent\textbf{Ground Truth Solution Filtering}
Regarding ground truth solutions, we collect five correct code submissions for each problem from a reliable online source. These submissions are subjected to a cleaning process to eliminate any potentially contaminated or duplicate code. This ensures the final ground truth solutions is robust and representative. 

Overall, the annotation process emphasizes a balance between automation and expert evaluation. By combining model-generated outputs with detailed human review, we aim to produce high-quality annotations that serve as a solid foundation for subsequent analyses. The multi-step approach not only ensures reliability but also promotes transparency in our methodology.

\section{Implementation Details}
\label{detailsimp}

We conduct all our experiments using a single Nvidia RTX A100 GPU for the 7B size LLMs, two A100 GPUs for the 13B size LLMs, and four A100 GPUs for the 34B size LLMs. For these open-source LLMs, we utilize the Xinference framework. For all LLMs, we employ nucleus sampling with a temperature of 0.7 and a top-p value of 0.95, allowing for a maximum of 10 iterations per stage with human programmers. For the pass@k metrics, we calculate it using the macro average method.

\subsection{Implementation of Real Human Experiments}
\label{setforhuamn}
We involve five computer science graduate students collaboratively solving competitive programming problems with LLMs. Each human participant receives detailed instructions (cf. Appendix \ref{tab:human_debug_ins}) before commencing the experiment, and is permitted to utilize any resources, including our dataset and internet search, to cooperate with one LLM, ensuring a realistic debugging process. Each participant in the human-subject study received a compensation of \$150 for their participation. Following \citet{shi2024can,tian2024debugbench}, human interaction is limited to textual feedback identifying errors (syntactic and semantic, see Table \ref{error_taxonomy}) rather than direct code editing. This debugging process ends until a correct solution is produced or a maximum number of iterations 10 is reached. During this process, the specific challenges LLM would encounter are not known in advance, potentially leading to high human labor. Thus, to mitigate this labor, we focus human participation on the debugging stage using a subset of 300 randomly selected problems of varying difficulty from \ourdata~(100 for easy, 100 for middle and 100 for hard problems). For the LLM, we consider GPT-4-Turbo (due to its balance of strong performance and reasonable cost). It is allows to refine its solutions based on both compiler feedback and simulator feedback played by O1-Mini (We term this process as Automatic Debug), which reduces the human workload for bug identification.

To ensure a rigorous and comprehensive analysis of the generated code, we conducted a detailed post-experiment review process. In this stage, all generated code was meticulously analyzed, and any identified bugs were carefully annotated. The error annotation task focused on code generated by GPT-4 Turbo, leveraging the expertise of a team of five graduate students in computer science. All members of the team possess substantial experience in Olympic-level competitive programming, equipping them with the necessary skills to identify subtle and complex coding issues. For each problem, at least two annotators independently reviewed the generated code to ensure accuracy and consistency in error identification. The annotation process was guided by a detailed set of instructions, outlined in Table \ref{guidelines for human}. These guidelines provided step-by-step instructions on how to identify, categorize, and label errors. Specific categories of errors included logical flaws, syntax issues, edge case failures, inefficiencies, and implementation inconsistencies. The standardized approach ensured uniformity across the annotations, reducing subjectivity and enhancing reliability. Once the annotations were completed, any discrepancies between annotators were resolved through consensus discussions, ensuring that the final error labels accurately reflected the issues present in the code. Examples of annotated bugs, including descriptions and their corresponding fixes, are provided in Table \ref{bug_example}, illustrating the depth and clarity of the annotation process. The annotated error dataset forms a crucial part of this study and serves as a valuable resource for understanding common pitfalls in LLM-generated code. To promote transparency and support future research, this supplementary dataset will be publicly released alongside the primary dataset. By sharing this resource, we aim to facilitate further investigation into the strengths and limitations of code-generation models while fostering advancements in the field of AI-assisted programming.

\subsection{LLM Implementation}
\label{sec:model}
We tested a total of 13 different large language models. The details of the models considered in our study are described in Table \ref{models_list}.

\begin{table*}[!h]
\centering
\begin{tabular}{|p{0.4\linewidth}|p{0.12\linewidth}|p{0.28\linewidth}|}
    \hline
    \textbf{Model Name} & \textbf{Cutoff Date} & \textbf{Link} \\
    \hline
    Deepseek-coder-6.7b-instruct & 09/2023 & \href{https://huggingface.co/deepseek-ai/deepseek-coder-6.7b-instruct}{deepseek-coder-6.7b-instruct} \\ 
    \hline
    Deepseek-coder-33b-instruct & 09/2023 & \href{https://huggingface.co/deepseek-ai/deepseek-coder-33b-instruct}{deepseek-coder-33b-instruct} \\ 
    \hline
    CodeLlama-7b-Instruct & 01/2023 & \href{https://huggingface.co/codellama/CodeLlama-7b-Instruct-hf}{CodeLlama-7b-Instruct} \\ 
    \hline
    CodeLlama-13b-Instruct & 01/2023 & \href{https://huggingface.co/codellama/CodeLlama-13b-Instruct-hf}{CodeLlama-13b-Instruct} \\ 
    \hline
    CodeLlama-34b-Instruct & 01/2023 & \href{https://huggingface.co/codellama/CodeLlama-34b-Instruct-hf}{CodeLlama-34b-Instruct} \\ 
    \hline
    Qwen2.5-Coder-7B & 06/2024 & \href{https://huggingface.co/Qwen/Qwen2.5-Coder-7B}{Qwen2.5-Coder-7B} \\ 
    \hline
    Qwen2.5-Coder-14B & 06/2024 & \href{https://huggingface.co/Qwen/Qwen2.5-Coder-14B}{Qwen2.5-Coder-14B} \\ 
    \hline
    Qwen2.5-Coder-32B & 06/2024 & \href{https://huggingface.co/Qwen/Qwen2.5-Coder-32B}{Qwen2.5-Coder-32B} \\ 
    \hline
    GPT-4-Turbo & 05/2023 & \href{https://openai.com/blog/new-models-and-developer-products-announced-at-devday}{GPT-4-Turbo} \\ 
    \hline
    GPT-4o & 11/2023 & \href{https://openai.com/index/spring-update}{GPT-4o} \\ 
    \hline
    Claude-3.5-Sonnet & 03/2024 & \href{https://www.anthropic.com/claude}{Claude-3.5-Sonnet} \\ 
    \hline
    Gemini-1.5-Pro & 11/2023 & \href{https://blog.google/technology/ai/gemini-api-developers-cloud}{Gemini-1.5-Pro} \\ 
    \hline
    O1-Mini & 12/2023 & \href{https://openai.com/o1/}{O1-Mini}\\
    \hline
\end{tabular}
\caption{LLMs Overview}
\label{models_list}
\end{table*}

\subsection{Implementation of Evaluation}

In the comprehension stage, after the LLM has developed an understanding of the problem statement, O1-Mini is used for automatic fact-checking against the annotations of the clarifications and requirements specified in the dataset, the detailed results are presented in Table \ref{tab:comprehension_res}. During the planning stage, the LLM selects a suitable algorithm, generates reasoning, and provides pseudocode. O1-Mini is then utilized for automatic fact-checking against the annotated summaries of the algorithms in the dataset. Since a single problem may correspond to multiple algorithm options, it is sufficient for the LLM to accurately select one of the algorithms (drawing on established fact-checking metrics \cite{das2023state}), the detailed results are presented in Table \ref{tab:planning_res}. In the coding stage, the LLM produces the complete code, which is subsequently tested using the test case examples provided in the problem statement through a compiler. In the debugging stage, all test cases are re-evaluated through a compiler based on the modified code from the LLM.

\section{Additional Results}
\label{more}
\subsection{Results of Pass@K}
Tables \ref{tab:overall_results_pass@3} and Table \ref{tab:overall_results_pass@5} present additional results for human-LLM competitive programming using Pass@3 and Pass@5 metrics, respectively. Key observations we can draw from these tables are in line with ones in the main body of this paper. 

\subsection{Case Studies}
For better understanding, Appendix \ref{cased} provides a case study illustrating how human feedback from simulator improves LLM performance in competitive programming. Additionally, Appendix \ref{errrrr} provides a case of LLM incorparating real human feedback to solve the coding problem. More examples are available in our dataset. 

\begin{table}[ht]
\centering
\resizebox{0.95\linewidth}{!}{ 
\begin{tabular}{c|c|c|c}
\hline
 Comprehension  Stage & Easy & Middle  & Hard\\   
\hline
O1-Mini & 0.99 &	0.99&	0.99\\
GPT-4o  &0.99	& 0.96	& 0.96  \\
GPT-4-Turbo  &0.98	&0.93	&0.91 \\
Gemini-1.5-Pro &0.98	&0.96	&0.95   \\
Claude-3.5-Sonnet &0.99	&0.96	&0.95  \\
Deepseek-coder-6.7b-instruct  &0.95	&0.89 	&0.84   \\
CodeLlama-7b-Instruct &0.93	&0.87	&0.82   \\
Qwen2.5-Coder-7B &0.94	&0.89 	&0.85  \\
CodeLlama-13b-Instruct  &0.94	&0.91	&0.86   \\
Qwen2.5-Coder-14B  &0.95	&0.93	&0.87   \\
Deepseek-coder-33b-instruct &0.96	&0.94	&0.90    \\
CodeLlama-34b-Instruct &0.94	&0.93	&0.88   \\
Qwen2.5-Coder-32B  &0.97	&0.95	&0.92   \\
\hline
\end{tabular}
}
\caption{Different LLMs' performance at comprehension stage}
\label{tab:comprehension_res}
\end{table}

\begin{table}[ht]
\centering
\resizebox{0.95\linewidth}{!}{ 
\begin{tabular}{c|c|c|c}
\hline
 Planning Stage & Easy & Middle  & Hard\\   
\hline
O1-Mini &0.97	&0.93	&0.90   \\
GPT-4o  &0.91	&0.77	&0.66   \\
GPT-4-Turbo  &0.82	&0.65	&0.48  \\
Gemini-1.5-Pro &0.91	&0.78	&0.64   \\
Claude-3.5-Sonnet &0.92 	&0.75	&0.62   \\
Deepseek-coder-6.7b-instruct  &0.51	&0.29 	&0.17  \\
CodeLlama-7b-Instruct &0.48	&0.25	&0.14   \\
Qwen2.5-Coder-7B &0.54	&0.33 	&0.19  \\
CodeLlama-13b-Instruct  &0.53	&0.31	&0.18   \\
Qwen2.5-Coder-14B  &0.66	&0.49	&0.24   \\
Deepseek-coder-33b-instruct &0.72	&0.50 	&0.39    \\
CodeLlama-34b-Instruct &0.64 	&0.37 	&0.28  \\
Qwen2.5-Coder-32B  &0.78	&0.57 	&0.43   \\
\hline
\end{tabular}
}
\caption{Different LLMs' performance at planning stage}
\label{tab:planning_res}
\end{table}

\subsection{Nuanced Understanding of Each Stage}
\label{nuance}
Our evaluation approach provides a reliable and objective framework for assessing LLMs in competitive programming by leveraging an expert-verified dataset to quantify understanding (accuracy of problem comprehension), planning (correct algorithm selection), ensure consistency (eliminating subjective human evaluation), and provide scalability (enabling rapid evaluation across numerous problems). Beyond this, to enable a more nuanced analysis of different LLMs at the comprehension and planning stages, we conducted detailed manual evaluations with the goal of gaining a deeper understanding of their performance. For each stage, we randomly selected 20 algorithmic problems(10 easy, 10 middle) for testing. Two computer science master’s students participated in this manual evaluation. All procedures were conducted under Institutional Review Board (IRB) approval.

\subsubsection{Analysis on Comprehension Stage}
We design four dimensions for in-depth evaluation: 
\begin{itemize}
    \item \textbf{Understanding of Problem Requirements:} Measures whether the model accurately identifies and interprets the core requirements of the problem, ensuring that the comprehension of the task aligns with the given problem statements.
    \item \textbf{Correctness of Identified Specifications:} Assesses whether the model accurately identifies any constraints or specifications inherent in the problem (e.g., time, space, or other domain-specific constraints).
    \item \textbf{Clarity of Functionality:} evaluates how clearly the model defines the purpose and behavior of the proposed solution. It measures whether the model communicates the solution's functionality in a concise and understandable way. 
\end{itemize}

As Table \ref{tab:nuanced_comprehension} shown, the results from the manual evaluation provide a more comprehensive display of different LLMs' performance at comprehension stage. The experimental results reveal that while all models exhibit high accuracy in requirements understanding and specification identification, small-scale open-source models show notable deficiencies in problem breakdown and functionality clarification compared to other models. These shortcomings are aspects that cannot be captured through automatic evaluation.

\subsubsection{Analysis on Planning Stage}
We designed four dimensions for in-depth evaluation: 
\begin{itemize}
    \item \textbf{Algorithm Selection Accuracy:} Whether the algorithm chosen by the model aligns with the problem’s requirements.
    \item \textbf{Pseudocode Generation Quality:} The correctness and completeness of the generated algorithm pseudocode.
    \item \textbf{Algorithm Justification Reasonableness:} Whether the rationale behind choosing the algorithm is sufficient and logical.
\end{itemize}

As Table \ref{tab:nuanced_planning} shown, the results from the manual evaluation provide a more comprehensive display of different LLMs' performance at planning stage. They further highlight that closed-source models (especially O1-mini) outperform open-source models in the planning stage, particularly in complexity analysis and algorithm justification —areas that cannot be assessed through automatic evaluation.

\begin{table}[htbp]
\centering
\resizebox{0.95\linewidth}{!}{
\begin{tabular}{lccc}
\toprule
\textbf{Coding Stage} & \textbf{Original} & \textbf{Automatic Feedback} & \textbf{Human Feedback} \\
\midrule
Easy & 13 & 15 & 16 \\
Middle & 5 & 6 & 8 \\
Hard & 2 & 4 & 5 \\
\midrule
\textbf{Total} & 20 & 25 & 29 \\
\bottomrule
\end{tabular}}
\caption{Performance in collaborating with real humans on coding tasks}
\label{tab:real_human_res}
\end{table}

\subsection{Collaborating with Real Human on Coding Stage}
\label{codinghuman}

To further investigate Human-LLM collaboration, we conducted experiments incorporating real human participants during the coding stage, utilizing the GPT-4-turbo model. Consistent with our human experiments in the debugging stage, five computer science master's students participated, and all procedures were conducted under Institutional Review Board (IRB) approval. And We employ O1-mini as the teacher programmer to provide automatic feedback. Due to time constraints, we employed a randomly selected set of 60 problems from an unseen dataset, with 20 problems each from the easy, middle, and hard categories. This dataset was used to evaluate the performance of the GPT-4-turbo model. Furthermore, to facilitate Human-LLM collaboration, we augmented the automatic feedback generated by our teacher-programmer simulator with additional human feedback.

\textbf{Experimental Results}. As shown in Table \ref{tab:real_human_res}, GPT-4-turbo initially solved 13 easy problems, 5 medium problems, and 2 hard problems. After receiving automatic feedback, the model was able to solve 15 easy problems, 6 medium problems, and 4 hard problems. With the addition of human feedback, it solved 16 easy problems, 8 medium problems, and 5 hard problems. These results highlight the effectiveness of Human-LLM collaboration in solution provision during the coding stage. However, compared to collaboration during the debugging stage, the improvement is relatively modest. This can be attributed to the teacher-programmer simulator, which, by already referencing the ground truth solution, offers the correct approach. Consequently, adding human feedback to an already accurate solution provides limited additional benefit.

\subsection{Analysis of Feedback from Human Simulators and Real Humans}
\subsubsection{Analysis on Human Simulators}
In this paper, we include two participant groups representing a range of programming expertise: student programmers and teacher programmers. The differences between these groups are as follows:

\noindent\textbf{Resource Utilization}. In the benchmark experiments, the O1-Mini model serves as the student programmer, offering feedback based on its internal knowledge. In contrast, the teacher programmer simulator leverages the full ELABORATIONSET dataset to guide the O1-Mini model, ensuring expert-level performance. This difference in resource utilization directly impacts the effectiveness of the feedback provided.

\noindent\textbf{Feedback Quality}. As shown in Table 5, the effectiveness of feedback is directly correlated to its specificity, detail, and strategic insights. Teacher feedback, by providing these elements, promotes a deeper and more effective learning experience for the LLM. Student feedback, lacking these features, is often less impactful, resulting in slower or less significant improvements in LLM performance. The key is that more precise and insightful feedback leads to better understanding and problem-solving capabilities in the LLM.

\subsubsection{Analysis on Real Humans}
\label{this}

As shown in table \ref{Debug_score} and \ref{Bug_Type}, human debugging demonstrates superior effectiveness compared to automatic debugging. In this module, we will discuss the difference between real human feedback and human simulator feedback at debugging stage.  We illustrate an example of human-LLM collaboration during the debugging stage in Appendix \ref{guidelines for human}. The effectiveness of feedback is directly tied to the accuracy of error identification and the level of detail in modification suggestions. Real human debugging excels in these aspects, providing more precise bug identification and more detailed modification suggestions compared to automatic debugging. In contrast, automatic feedback, lacking these critical features, often proves less impactful, resulting in slower or less substantial debugging improvements in LLM performance. Ultimately, the key to enhancing the LLM's bug-fixing capabilities lies in the ability to deliver accurate bug identification and detailed modification suggestions.

\clearpage

\begin{table*}[ht]
\centering
\resizebox{0.95\linewidth}{!}{ 
\begin{tabular}{c|c|c|c|c|c}
\hline
& \multicolumn{1}{c|}{Problem Requirements} & \multicolumn{1}{c|}{Specifications Identification} & \multicolumn{1}{c|}{Problem Breakdown} & \multicolumn{1}{c|}{Functionality} & \multicolumn{1}{c}{All} \\
 & \multicolumn{1}{c|}{Understanding} & \multicolumn{1}{c|}{Correctness} & \multicolumn{1}{c|}{Correctness} & \multicolumn{1}{c|}{Clarification} &  \multicolumn{1}{c}{Correct}\\
\hline
O1-Mini & 20 & 20 & 20 & 20 & 20 \\
GPT-4o  & 20 & 20 & 19 & 19 & 19 \\
GPT-4-Turbo  & 20 & 19 & 18 & 18 & 18 \\
Gemini-1.5-Pro & 20 & 20 & 19 & 19 &19  \\
Claude-3.5-Sonnet & 20 & 20 & 19 & 19 & 19 \\ 

Deepseek-coder-6.7b-instruct  & 19 & 17 & 16 & 18 & 16 \\
CodeLlama-7b-Instruct & 17 & 17 & 15 & 17  & 15 \\
Qwen2.5-Coder-7B & 19 & 19 & 18 & 18 & 18\\
CodeLlama-13b-Instruct  & 19 & 19 & 17 & 17 & 17\\
Qwen2.5-Coder-14B  & 20 & 19 & 18 & 19 & 18 \\
Deepseek-coder-33b-instruct & 20 & 20 & 19 & 19 & 19 \\
CodeLlama-34b-Instruct & 20 & 20 & 18 & 18 & 18 \\
Qwen2.5-Coder-32B  & 20 & 20 & 20 & 20 & 20\\
\midrule
\textit{\textbf{Average Performance}}  & 19.5 &	19.2 & 	18.2 &	18.5 &	18.2
 \\
\hline
\end{tabular}
}
\caption{Nuanced evaluation for different LLMs' performance at comprehension stage}
\label{tab:nuanced_comprehension}
\end{table*}

\begin{table*}[ht]
\centering
\resizebox{0.95\linewidth}{!}{ 
\begin{tabular}{c|c|c|c|c|c}
\hline
& \multicolumn{1}{c|}{Algorithm Selection} & \multicolumn{1}{c|}{Pseudocode Generation} & \multicolumn{1}{c|}{Complexity Analysis} & \multicolumn{1}{c|}{Algorithm Justification} & \multicolumn{1}{c}{All} \\
 & \multicolumn{1}{c|}{Accuracy} & \multicolumn{1}{c|}{Quality} & \multicolumn{1}{c|}{Accuracy} & \multicolumn{1}{c|}{Reasonableness} &  \multicolumn{1}{c}{Correct}\\
\hline
O1-Mini & 20 & 20 & 19 & 19 & 19 \\
GPT-4o  & 19 & 18 & 17 & 18 & 17 \\
GPT-4-Turbo  & 18 & 17 & 16 & 15 & 15 \\
Gemini-1.5-Pro & 19 & 18 & 18 & 17 & 17 \\
Claude-3.5-Sonnet & 19 & 18 & 17 & 18 & 17 \\

Deepseek-coder-6.7b-instruct  & 10 & 10 & 8 & 7 & 7 \\
CodeLlama-7b-Instruct & 7 & 6 & 5 & 5  & 5 \\
Qwen2.5-Coder-7B & 10 & 10 & 9 & 8 & 8\\
CodeLlama-13b-Instruct  & 11 & 9 & 9 & 8 & 8\\
Qwen2.5-Coder-14B  & 13 & 12 & 12 & 13 & 12 \\
Deepseek-coder-33b-instruct & 16 & 14 & 13 & 14 & 13 \\
CodeLlama-34b-Instruct & 13 & 11 & 10 & 11 & 10 \\
Qwen2.5-Coder-32B  & 17 & 16 & 15 & 14 & 14\\
\midrule
\textit{\textbf{Average Performance}}   & 14.8 &	13.8 &	12.9 &	12.8 &	12.5 \\
\hline
\end{tabular}
}
\caption{Nuanced evaluation for different LLMs' performance at planning stage}
\label{tab:nuanced_planning}
\end{table*}





\begin{table*}[t]
\small
\centering
\resizebox{\textwidth}{!}{
\begin{tabular}{@{}l|cccc|cccc@{}}
\toprule
\multirow{2}{*}{\textbf{Model} (Cut-off Date|Release Date)}
& \multicolumn{4}{c|}{\textbf{Contamination Evaluation}} &\multicolumn{4}{c}{\textbf{Contamination-free Evaluation}} \\ \cline{2-9}
& \textbf{Easy} & \textbf{Middle} & \textbf{Hard} & \textbf{Overall} & \textbf{Easy} & \textbf{Middle} & \textbf{Hard} & \textbf{Overall}  \\ \midrule

\textbf{O1-Mini} (2023-12 | 2024-09) & 93.5  & 78.0 & 50.7 & 74.1  & 86.2 &73.2  &34.8  & 64.7 \\
\midrule

\textbf{GPT-4o} (2023-11 | 2024-05) &84.4  &54.5  &23.2  &54.0  &79.0  &34.8  &11.5  &41.8 \\
\quad + Student Programmer Feedback &87.7  &57.5  &27.2  &57.5  &82.3  &38.5  &16.4  &45.7 \\ 
\quad + Teacher Programmer Feedback &92.8  &72.1  &42.8  &69.2  &86.5  &47.5  &26.1  &53.4 \\  \cline{2-9}

\textbf{GPT-4-Turbo} (2023-05 | 2023-11) &74.0  &44.1  &9.7   &42.6  &70.3  &30.2  &6.5   &35.7 \\
\quad + Student Programmer Feedback  &80.2  &50.6  &13.5  &48.1  &76.4  &36.5  &9.8   &40.9 \\ 
\quad + Teacher Programmer Feedback  &89.9  &64.7  &22.5  &59.0  &81.3  &43.8  &16.0  &47.0 \\  \cline{2-9}

\textbf{Gemini-1.5-pro} (2023-11 | 2024-02)  &85.3  &52.2  &24.6  &54.0  &79.1  &36.5  &10.4  &42.0 \\
\quad + Student Programmer Feedback  &88.7  &54.6  &28.1  &57.1  &81.6  &38.9  &14.7  &45.1 \\ 
\quad + Teacher Programmer Feedback  &96.0  &71.5  &41.0  &69.5  &87.5  &44.5  &27.1  &53.0 \\  \cline{2-9}

\textbf{Claude-3.5} (2024-03 | 2024-06)  &82.1  &55.9  &18.3  &52.1  &80.5  &37.8  &6.1   &41.5 \\
\quad + Student Programmer Feedback  &86.8  &60.5  &27.0  &58.1  &82.7  &40.8  &8.8   &44.1 \\ 
\quad + Teacher Programmer Feedback  &93.9  &73.3  &37.4  &68.2  &89.8  &48.6  &18.5  &52.3 \\  \cline{2-9}

\textbf{\textit{Avg.}}  &81.5  &51.7  &19.0  &50.7  &77.2  &34.8  &8.6  &40.3 \\ 
\quad + Student Programmer Feedback  &85.9 (+4.4)  &55.8 (+4.1)  &24.0 (+5.0)  &55.2 (+4.5)  &80.8 (+3.6)  &38.7 (+3.9)  &12.4 (+3.8)  &44.0 (+3.7) \\ 
\quad + Teacher Programmer Feedback  &93.2 (+11.7)  &70.4 (+18.7)  &35.9 (+16.9)  &66.5 (+15.8)  &86.3 (+9.1)  &46.1 (+11.3)  &21.9 (+13.3)  &51.4 (+11.1) \\ \midrule

\multicolumn{9}{c}{\textbf{$\sim$7B Scale}} \\ \midrule

\textbf{CodeLlama-7B} (2023-01 | 2024-01)  &32.8  &6.5  &0.6  &13.3  &16.2  &2.3  &0.3  &6.3 \\
\quad + Student Programmer Feedback  &39.6  &11.2  &2.5  &17.8  &26.1  &3.4  &1.6  &10.4 \\ 
\quad + Teacher Programmer Feedback  &52.4  &19.3  &7.7  &26.5  &38.8  &9.3  &5.3  &17.8 \\  \cline{2-9}

\textbf{Deepseek-Coder-6.7B} (2023-09 | 2023-11)  &43.8  &16.9  &2.0  &20.9  &23.1  &7.7  &0.8  &10.5 \\
\quad + Student Programmer Feedback  &50.0  &20.5  &4.8  &25.1  &30.2  &12.3  &2.3  &14.9 \\ 
\quad + Teacher Programmer Feedback  &63.1  &30.6  &9.1  &34.3  &42.6  &26.4  &6.9  &25.3 \\  \cline{2-9}

\textbf{Qwen2.5-Coder-7B} (2024-06 | 2024-11)  &65.7  &24.6  &5.4  &31.9  &52.5  &10.3  &0.6  &21.1 \\
\quad + Student Programmer Feedback  &75.7  &29.3  &7.5  &37.5  &58.3  &13.5  &2.6  &24.8 \\ 
\quad + Teacher Programmer Feedback  &82.4  &39.1  &12.3  &44.6  &62.5  &23.8  &6.6  &31.0 \\  \cline{2-9}

\textbf{\textit{Avg.}}  &47.4  &16.0  &2.7  &22.0  &30.6  &6.8  &0.6  &12.6 \\
\quad + Student Programmer Feedback  &55.1 (+7.7)  &20.3 (+4.3)  &4.9 (+2.2)  &26.8 (+4.8)  &38.2 (+7.6)  &9.7 (+2.9)  &2.2 (+1.6)  &16.7 (+4.1) \\ 
\quad + Teacher Programmer Feedback  &66.0 (+18.6)  &29.7 (+13.7)  &9.7 (+7.0)  &35.1 (+13.1)  &48.0 (+17.4)  &19.8 (+13.0)  &6.3 (+5.7)  &24.7 (+12.1) \\ \midrule

\multicolumn{9}{c}{\textbf{$\sim$13B Scale}} \\ \midrule

\textbf{CodeLlama-13B} (2023-01 | 2024-01)  &38.5  &8.0  &1.9  &16.1  &25.0  &3.3  &0.3  &9.5 \\
\quad + Student Programmer Feedback  &44.3  &13.1  &3.2  &20.2  &28.1  &10.7  &1.6  &13.5 \\ 
\quad + Teacher Programmer Feedback  &47.7  &21.9  &6.5  &25.4  &32.3  &16.1  &3.5  &17.3 \\  \cline{2-9}

\textbf{Qwen2.5-Coder-14B} (2024-06 | 2024-11)  &75.4  &31.5  &8.6  &38.5  &62.3  &16.6  &2.5  &27.1 \\
\quad + Student Programmer Feedback  &81.0  &36.9  &11.4  &43.1  &66.4  &20.8  &4.6  &30.6 \\ 
\quad + Teacher Programmer Feedback  &86.5  &45.6  &15.9  &49.3  &71.6  &26.7  &7.6  &35.3 \\  \cline{2-9}

\textbf{\textit{Avg.}}  &57.0  &19.8  &5.3  &27.3  &43.7  &10.0  &1.4  &18.3 \\
\quad + Student Programmer Feedback  &62.7 (+5.7)  &25.0 (+5.2)  &7.3 (+2.0)  &31.7 (+4.4)  &47.3 (+3.6)  &15.8 (+5.8)  &3.1 (+1.7)  &22.1 (+3.8) \\ 
\quad + Teacher Programmer Feedback  &67.1 (+10.1)  &33.8 (+14.0)  &11.2 (+5.9)  &37.4 (+10.1)  &52.0 (+8.3)  &21.4 (+11.4)  &5.6 (+4.2)  &26.3 (+8.0) \\ \midrule

\multicolumn{9}{c}{\textbf{$\sim$34B Scale}} \\ \midrule

\textbf{CodeLlama-34B} (2023-01 | 2024-01)  &40.3  &8.6  &3.5  &17.5  &26.7  &5.6  &1.1  &11.1 \\
\quad + Student Programmer Feedback  &45.4  &13.5  &4.5  &21.1  &28.2  &9.2  &2.6  &13.3 \\ 
\quad + Teacher Programmer Feedback  &52.7  &20.7  &6.9  &26.8  &34.7  &14.2  &5.0  &18.0 \\  \cline{2-9}

\textbf{Deepseek-Coder-33B} (2023-09 | 2023-11)  &68.3  &25.8  &4.8  &33.0  &54.0  &11.4  &1.4  &22.3 \\
\quad + Student Programmer Feedback  &80.1  &31.6  &7.8  &39.8  &59.9  &14.5  &3.5  &26.0 \\ 
\quad + Teacher Programmer Feedback  &85.2  &44.1  &13.8  &47.7  &73.4  &22.4  &6.2  &34.0 \\  \cline{2-9}

\textbf{Qwen2.5-Coder-32B} (2024-06 | 2024-11)  &82.0  &45.2  &10.1  &45.8  &74.6  &22.3  &3.6  &33.5 \\
\quad + Student Programmer Feedback  &86.1  &49.3  &12.3  &49.2  &77.1  &25.4  &4.5  &35.7 \\ 
\quad + Teacher Programmer Feedback  &91.4  &58.7  &17.7  &55.9  &82.0  &33.1  &8.5  &41.2 \\  \cline{2-9}

\textbf{\textit{Avg.}}  &63.5  &26.5  &6.1  &32.1  &51.8  &13.1  &2.0  &22.3 \\
\quad + Student Programmer Feedback  &70.5 (+7.0)  &31.5 (+5.0)  &8.2 (+2.1)  &36.7 (+4.6)  &55.1 (+3.3)  &16.4 (+3.3)  &3.5 (+1.5)  &25.0 (+2.7) \\ 
\quad + Teacher Programmer Feedback  &76.4 (+12.9)  &41.2 (+14.7)  &12.8 (+6.7)  &43.5 (+11.4)  &63.4 (+11.6)  &23.2 (+10.1)  &6.6 (+4.6)  &31.1 (+8.8) \\ \midrule

\textbf{\textit{Average over All LLMs}} &64.4  &31.2  &9.4  &35.0  &53.6  &18.2  &3.8  &25.2 \\
\quad + Student Programmer Feedback &70.5 (+6.1)  &35.7 (+4.5)  &12.5 (+3.1)  &39.6 (+4.6)  &58.1 (+4.5)  &22.0 (+3.8)  &6.1 (+2.3)  &28.8 (+3.6) \\ 
\quad + Teacher Programmer Feedback &77.8 (+13.4)  &46.8 (+15.6)  &19.5 (+10.1)  &48.0 (+13.0)  &65.3 (+11.7)  &29.7 (+11.5)  &11.4 (+7.6)  &35.5 (+10.3) \\

 \bottomrule
\end{tabular}} 
\caption{\ours~benchmark analysis of human-LLM competitive programming across different LLM backbones and varying levels of human feedback expertise. Because O1-Mini was recently released, our experiments with this model have been deferred. Here, we report Pass@3 scores (\%).}
\label{tab:overall_results_pass@3}
\vspace{-3mm}
\end{table*}

\begin{table*}[t]
\small
\centering
\resizebox{\textwidth}{!}{
\begin{tabular}{@{}l|cccc|cccc@{}}
\toprule
\multirow{2}{*}{\textbf{Model} (Cut-off Date|Release Date)}
& \multicolumn{4}{c|}{\textbf{Contamination Evaluation}} &\multicolumn{4}{c}{\textbf{Contamination-free Evaluation}} \\ \cline{2-9}
& \textbf{Easy} & \textbf{Middle} & \textbf{Hard} & \textbf{Overall} & \textbf{Easy} & \textbf{Middle} & \textbf{Hard} & \textbf{Overall}  \\ \midrule

\textbf{O1-Mini} (2023-12 | 2024-09) & 95.3  & 81.4  & 52.8  & 76.5  & 89.0 & 76.4 & 37.5 & 67.6 \\
\midrule

\textbf{GPT-4o} (2023-11 | 2024-05) &86.7  &57.8  &24.9  &56.5  &82.2  &36.8  &12.8  &43.9 \\
\quad + Student Programmer Feedback &89.0  &61.0  &29.0  &59.7  &85.6  &40.4  &18.0  &48.0 \\ 
\quad + Teacher Programmer Feedback &95.0  &76.3  &45.1  &72.1  &89.1  &50.3  &28.1  &55.8 \\  \cline{2-9}

\textbf{GPT-4-Turbo} (2023-05 | 2023-11) &75.7  &46.8  &10.6  &44.4  &72.0  &32.4  &7.2  &37.2 \\
\quad + Student Programmer Feedback  &82.0  &53.5  &14.6  &50.0  &78.5  &38.5  &10.7  &42.6 \\ 
\quad + Teacher Programmer Feedback  &92.1  &68.6  &24.0  &61.6  &84.2  &46.3  &17.3  &49.3 \\  \cline{2-9}

\textbf{Gemini-1.5-pro} (2023-11 | 2024-02) &87.0  &55.5  &26.1  &56.2  &82.2  &39.3  &11.7  &44.4 \\
\quad + Student Programmer Feedback  &91.0  &58.5  &30.1  &59.9  &85.2  &42.7  &16.0  &48.0 \\ 
\quad + Teacher Programmer Feedback  &99.8  &75.7  &44.7  &73.4  &91.0  &47.8  &29.3  &56.0 \\  \cline{2-9}

\textbf{Claude-3.5} (2024-03 | 2024-06)  &84.0  &59.3  &19.7  &54.3  &83.7  &40.5  &6.9  &43.7 \\
\quad + Student Programmer Feedback  &89.3  &64.2  &29.4  &61.0  &86.2  &43.8  &9.6  &46.5 \\ 
\quad + Teacher Programmer Feedback  &97.2  &77.7  &40.3  &71.7  &93.3  &51.5  &20.2  &55.0 \\  \cline{2-9}

\textbf{\textit{Avg.}}  &83.4  &54.9  &20.3  &52.9  &80.0  &37.3  &9.7  &42.3 \\ 
\quad + Student Programmer Feedback  &87.8 (+4.4)  &59.3 (+4.4)  &25.8 (+5.5)  &57.7 (+4.8)  &83.9 (+3.9)  &41.4 (+4.1)  &13.6 (+3.9)  &46.3 (+4.0) \\ 
\quad + Teacher Programmer Feedback  &96.0 (+12.6)  &74.6 (+19.7)  &38.5 (+18.2)  &69.7 (+16.8)  &89.4 (+9.4)  &49.0 (+11.7)  &23.7 (+14.0)  &54.0 (+11.7) \\ \midrule

\multicolumn{9}{c}{\textbf{$\sim$7B Scale}} \\ \midrule

\textbf{CodeLlama-7B} (2023-01 | 2024-01)  &33.9  &6.8  &0.6  &13.8  &16.7  &2.4  &0.3  &6.5 \\
\quad + Student Programmer Feedback  &40.9  &11.6  &2.7  &18.4  &26.9  &3.6  &1.7  &10.7 \\ 
\quad + Teacher Programmer Feedback  &54.2  &20.4  &8.2  &27.6  &40.3  &9.8  &5.7  &18.6 \\  \cline{2-9}

\textbf{Deepseek-Coder-6.7B} (2023-09 | 2023-11) &45.0  &17.7  &2.1  &21.6  &24.0  &8.1  &0.8  &11.0 \\
\quad + Student Programmer Feedback  &52.0  &21.5  &5.1  &26.2  &31.3  &13.0  &2.5  &15.6 \\ 
\quad + Teacher Programmer Feedback  &65.5  &32.3  &9.7  &35.8  &44.0  &27.9  &7.4  &26.4 \\  \cline{2-9}

\textbf{Qwen2.5-Coder-7B} (2024-06 | 2024-11) &67.9  &25.8  &5.8  &33.2  &54.6  &10.9  &0.7  &22.1 \\
\quad + Student Programmer Feedback  &78.3  &30.8  &8.1  &39.1  &60.1  &14.3  &2.8  &25.7 \\ 
\quad + Teacher Programmer Feedback  &85.1  &41.4  &13.4  &46.6  &64.9  &25.2  &7.1  &32.4 \\  \cline{2-9}

\textbf{\textit{Avg.}}  &48.9  &16.8  &2.8  &22.9  &31.8  &7.1  &0.6  &13.2 \\
\quad + Student Programmer Feedback  &57.1 (+8.2)  &21.3 (+4.5)  &5.3 (+2.5)  &27.9 (+5.0)  &39.4 (+7.6)  &10.3 (+3.2)  &2.3 (+1.7)  &17.3 (+4.1) \\ 
\quad + Teacher Programmer Feedback  &68.3 (+19.4)  &31.4 (+14.6)  &10.4 (+7.6)  &36.7 (+13.8)  &49.7 (+17.9)  &21.0 (+13.9)  &6.7 (+6.1)  &25.8 (+12.6) \\ \midrule

\multicolumn{9}{c}{\textbf{$\sim$13B Scale}} \\ \midrule

\textbf{CodeLlama-13B} (2023-01 | 2024-01)  &39.8  &8.5  &2.0  &16.8  &26.0  &3.5  &0.3  &9.9 \\
\quad + Student Programmer Feedback  &46.1  &13.9  &3.4  &21.1  &29.2  &11.3  &1.7  &14.1 \\ 
\quad + Teacher Programmer Feedback  &49.6  &23.2  &6.9  &26.6  &33.4  &17.0  &3.8  &18.1 \\  \cline{2-9}

\textbf{Qwen2.5-Coder-14B} (2024-06 | 2024-11)  &78.0  &33.4  &9.1  &40.2  &64.2  &17.6  &2.7  &28.2 \\
\quad + Student Programmer Feedback  &83.5  &39.2  &12.1  &44.9  &68.3  &22.0  &4.9  &31.7 \\ 
\quad + Teacher Programmer Feedback  &89.1  &48.3  &16.8  &51.4  &73.1  &28.3  &8.2  &36.5 \\  \cline{2-9}

\textbf{\textit{Avg.}}  &58.9  &21.0  &5.6  &28.5  &45.1  &10.6  &1.5  &19.1 \\
\quad + Student Programmer Feedback  &64.8 (+5.9)  &26.6 (+5.6)  &7.8 (+2.2)  &33.0 (+4.5)  &48.8 (+3.7)  &16.7 (+6.1)  &3.3 (+1.8)  &22.9 (+3.8) \\ 
\quad + Teacher Programmer Feedback  &69.4 (+10.5)  &35.8 (+14.8)  &11.9 (+6.3)  &39.0 (+10.5)  &53.3 (+8.2)  &22.7 (+12.1)  &6.0 (+4.5)  &27.3 (+8.2) \\ \midrule

\multicolumn{9}{c}{\textbf{$\sim$34B Scale}} \\ \midrule

\textbf{CodeLlama-34B} (2023-01 | 2024-01)  &41.5  &9.0  &3.8  &18.1  &27.6  &5.9  &1.2  &11.6 \\
\quad + Student Programmer Feedback  &46.8  &14.2  &4.9  &22.0  &29.3  &9.7  &2.8  &13.9 \\ 
\quad + Teacher Programmer Feedback  &54.1  &21.8  &7.4  &27.8  &36.0  &15.1  &5.4  &18.8 \\  \cline{2-9}

\textbf{Deepseek-Coder-33B } (2023-09 | 2023-11)  &70.1  &27.2  &5.2  &34.2  &55.9  &12.0  &1.6  &23.2 \\
\quad + Student Programmer Feedback  &83.2  &33.4  &8.4  &41.7  &62.1  &15.4  &3.8  &27.1 \\ 
\quad + Teacher Programmer Feedback  &88.7  &46.7  &15.0  &50.1  &75.8  &23.7  &6.8  &35.4 \\  \cline{2-9}

\textbf{Qwen2.5-Coder-32B } (2024-06 | 2024-11)  &84.5  &47.1  &11.0  &47.5  &76.9  &23.6  &3.9  &34.8 \\
\quad + Student Programmer Feedback  &89.0  &51.5  &13.3  &51.3  &79.2  &26.9  &4.8  &37.0 \\ 
\quad + Teacher Programmer Feedback  &94.3  &61.6  &19.0  &58.3  &84.6  &35.1  &9.2  &43.0 \\  \cline{2-9}

\textbf{\textit{Avg.}}  &65.4  &27.8  &6.7  &33.3  &53.5  &13.8  &2.2  &23.2 \\
\quad + Student Programmer Feedback  &73.0 (+7.6)  &33.0 (+5.2)  &8.9 (+2.2)  &38.3 (+5.0)  &56.9 (+3.4)  &17.3 (+3.5)  &3.8 (+1.6)  &26.0 (+2.8) \\ 
\quad + Teacher Programmer Feedback  &79.0 (+13.6)  &43.4 (+15.6)  &13.8 (+7.1)  &45.4 (+12.1)  &65.5 (+12.0)  &24.6 (+10.8)  &7.1 (+4.9)  &32.4 (+9.2) \\ \midrule

\textbf{\textit{Average over All LLMs}} &66.2  &32.9  &10.1  &36.4  &55.5  &19.4  &4.2  &26.4 \\
\quad + Student Programmer Feedback &72.6 (+6.4)  &37.8 (+4.9)  &13.4 (+3.3)  &41.3 (+4.9)  &60.2 (+4.7)  &23.5 (+4.1)  &6.6 (+2.4)  &30.1 (+3.7) \\ 
\quad + Teacher Programmer Feedback &80.4 (+14.2)  &49.5 (+16.6)  &20.9 (+10.8)  &50.3 (+13.9)  &67.5 (+12.0)  &31.5 (+12.1)  &12.4 (+8.2)  &37.1 (+10.7) \\

 \bottomrule
\end{tabular}} 
\caption{\ours~benchmark analysis of human-LLM competitive programming across different LLM backbones and varying levels of human feedback expertise. Because O1-Mini was recently released, our experiments with this model have been deferred. Here, we report Pass@5 scores (\%)}
\label{tab:overall_results_pass@5}
\vspace{-3mm}
\end{table*}

\clearpage

\begin{promptbox}[Random Input Generator Prompt]{softblue}
\label{Random_input}

You are an expert programmer and your goal is to generate test case input based based on the problem statement \{PROBLEM\_STATEMENT\}, input format \{INPUT\_FORMAT\}, output format \{OUTPUT\_FORMAT\}, test case example \{TEST\_CASE\_EXAMPLE\}, the input number lower boundary input\_min\{INPUT\_MIN\}, the input number upper\_boundary{INPUT\_MAX}. You will create appropriate generators and ultimately develop a test\_case\_inputs function that returns a list of varied inputs sampled from the generator. Be sure to adhere closely to the instructions and constraints outlined in the problem statement. 

\begin{verbatim}
import numpy as np

def random_input_generator(weight_min, weight_max, size_min, size_max):
    weights_size = np.random.randint(size_min, size_max + 1)
    weights = np.random.randint(weight_min, weight_max, size=weights_size).tolist()
    k = np.random.randint(1, len(weights) + 1)
    return weights, k

def construct_inputs():
    inputs_list = []
    
    for _ in range(15):
        inputs_list.append(random_input_generator(input\_min, input\_max ,1))
    
    return inputs_list


\end{verbatim}

Construct a random input generator. Return a single function named test\_case\_input that creates diverse inputs, following the format used in the example above.

\end{promptbox}

\clearpage

\begin{promptbox}[Algorithm konwledge summaries examples]{softblue}
\label{knowledge_summary_examples}

\textbf{Greedy: }  \\
A greedy algorithm is an approach for solving optimization problems by making a sequence of choices, each of which looks best at the moment. The idea is to choose the option that seems the most advantageous at each stage with the hope that these local optimal choices will lead to a global optimal solution. Greedy algorithms are typically used in problems like finding the minimum spanning tree, shortest paths in graphs, and various resource allocation problems. \\

The key characteristics of greedy algorithms are: \\

1. Locally Optimal Choice: At each step, the algorithm chooses the best option available without considering the global context. \\
2. Feasibility: The choice made must satisfy the problem's constraints. \\
3. Irrevocability: Once a choice is made, it cannot be undone. \\

Pseudocode for a Greedy Algorithm:  \\

\begin{verbatim}
Function GreedyFractionalKnapsack(capacity, items):
    // items is a list of tuples (value, weight)
    // Calculate value-to-weight ratio for each item
    for each item in items:
        item.ratio = item.value / item.weight

    // Sort items by their value-to-weight ratio in descending order
    Sort items by item.ratio in descending order

    total_value = 0
    for each item in items:
        if capacity == 0:
            break
        
        // Check if the item can be added fully
        if item.weight <= capacity:
            capacity -= item.weight
            total_value += item.value
        else:
            // Add the fraction of the remaining item
            total_value += item.value * (capacity / item.weight)
            capacity = 0 // Knapsack is full

    return total_value
\end{verbatim}

\textbf{Depth-First Search: } \\
Depth-First Search (DFS) is an algorithm for traversing or searching tree or graph data structures. The algorithm starts at the root (or an arbitrary node in a graph) and explores as far as possible along each branch before backtracking. This method is particularly useful for exploring all possible paths in a graph or tree, finding connected components, and solving puzzles with a single solution. \\

Key Characteristics of DFS: \\
1. Exploration Depth: DFS explores the deepest nodes first before backtracking. \\
2. Stack-Based: It can be implemented using recursion or an explicit stack data structure. \\
3. Complete: In finite graphs, DFS will visit all reachable nodes.
Space Complexity: The space complexity can be high due to the depth of recursion or the size of the stack. \\

\begin{verbatim}
Function DFS(graph, start_node):
    Initialize an empty set called visited
    Call DFS_Visit(graph, start_node, visited)

Function DFS_Visit(graph, node, visited):
    // Mark the current node as visited
    visited.add(node)
    Process the node (e.g., print or store the node value)
    
    // Recursively visit each adjacent node
    for each neighbor in graph[node]:
        if neighbor not in visited:
            DFS_Visit(graph, neighbor, visited)

\end{verbatim}

\end{promptbox}

\clearpage

\section{Human Feeback Case Study}
\label{cased}
We provided a detailed example showing how teacher programmer help Deepseek-Coder-33B solve competitive programming problems. \\

\begin{promptbox}[Codeforces Problem 1994 C]{softblue}
\label{tab:case_study_1}

\textbf{Problem Statement:}\\
Yaroslav is playing a computer game, and at one of the levels, he encountered \( n \) mushrooms arranged in a row. Each mushroom has its own level of toxicity; the \( i \)-th mushroom from the beginning has a toxicity level of \( a_i \). Yaroslav can choose two integers \( 1 \le l \le r \le n \), and then his character will take turns from left to right to eat mushrooms from this subsegment one by one, i.e., the mushrooms with numbers \( l, l+1, l+2, \ldots, r \).

The character has a toxicity level \( g \), initially equal to \( 0 \). The computer game is defined by the number \( x \) — the maximum toxicity level at any given time. When eating a mushroom with toxicity level \( k \), the following happens:
- The toxicity level of the character is increased by \( k \).
- If \( g \leq x \), the process continues; otherwise, \( g \) becomes zero and the process continues.

Yaroslav became interested in how many ways there are to choose the values of \( l \) and \( r \) such that the final value of \( g \) is not zero. Help Yaroslav find this number!
\\

\textbf{Input Format:}\\
Each test consists of multiple test cases. The first line contains an integer \( t \) (\( 1 \le t \le 10^4 \)) — the number of test cases. Then follows the description of the test cases.
- The first line of each test case contains two integers \( n \), \( x \) (\( 1 \leq n \leq 2 \cdot 10^5, 1 \le x \le 10^9 \)) — the number of mushrooms and the maximum toxicity level.
- The second line of each test case contains \( n \) numbers \( a_1, a_2, \ldots, a_n \) (\( 1 \leq a_i \leq 10^9 \)).

It is guaranteed that the sum of \( n \) over all test cases does not exceed \( 2 \cdot 10^5 \).
\\

\textbf{Output Format:} \\
For each test case, output a single number — the number of subsegments such that the final value of \( g \) will not be zero.
\\

\begin{verbatim}
Input
5
4 2
1 1 1 1
3 2
1 2 3
1 6
10
6 3
1 2 1 4 3 8
5 999999999
999999999 999999998 1000000000 1000000000 500000000

Output
8
2
0
10
7
\end{verbatim}

\end{promptbox}

\clearpage

\begin{promptbox}[Deepseek-Coder-33B generation at comprehension stage]{softblue}

\textbf{"Problem Requirements":}\\ 

1. "Input:" 
   - The first line contains an integer \( t \) (number of test cases).
   - Each test case consists of:
     - A line containing two integers \( n \) (the number of mushrooms) and \( x \) (the maximum toxicity level).
     - A line containing \( n \) integers \( a_1, a_2, \ldots, a_n \) (the toxicity levels of the mushrooms).

2. "Output:" 
   - For each test case, output a single integer representing the number of valid subsegments for which the final toxicity level \( g \) is not zero after consuming the mushrooms in that segment.

3. "Functionality:" 
   - The program should efficiently handle multiple test cases with constraints on the total number of mushrooms across all test cases.

\textbf{"Problem Specification":} 
- Constraints:
  - \( 1 \le t \le 10^4 \)
  - \( 1 \le n \le 2 \cdot 10^5 \)
  - \( 1 \le x \le 10^9 \)
  - \( 1 \le a_i \le 10^9 \)
  - The sum of \( n \) across all test cases does not exceed \( 2 \cdot 10^5 \).
  \\
- Each subsegment is defined by two indices \( l \) and \( r \) (1-based indexing) such that \( 1 \le l \le r \le n \).

\end{promptbox}

\begin{promptbox}[Teacher programmer feedback at comprehension stage]{softblue}

\textbf{Teacher Programmer Generation:} \\
"Unsatisfied": "You have not understood the problem statement, missing the requirements and specifications. Let me clarify the problem statement for you. Please modify your answer accordingly."\\

**Problem Requirements**:

1. **Input**:
    - The first line contains an integer \( t \) (\( 1 \leq t \leq 10^4 \)) — the number of test cases.
    - For each test case:
        - The first line contains two integers \( n \) and \( x \) (\( 1 \leq n \leq 2 \times 10^5 \), \( 1 \leq x \leq 10^9 \)) — the number of mushrooms and the maximum toxicity level.
        - The second line contains \( n \) integers \( a_1, a_2, \ldots, a_n \) (\( 1 \leq a_i \leq 10^9 \)) — the toxicity levels of the mushrooms.

2. **Output**:
    - For each test case, output a single integer — the number of subsegments \([l, r]\) (\( 1 \leq l \leq r \leq n \)) such that after processing the subsegment, the final value of toxicity level \( g \) is not zero.

3. **Functionality**:
    - For each subsegment \([l, r]\), simulate the process where the character starts with toxicity level \( g = 0 \).
    - The character eats mushrooms from left to right within the subsegment:
        - Upon eating the \( i \)-th mushroom (\( l \leq i \leq r \)):
            - Increase \( g \) by \( a_i \).
            - If \( g \leq x \), continue to the next mushroom.
            - If \( g > x \), reset \( g \) to zero and continue.
    - After eating all mushrooms in the subsegment, check if \( g \) is not zero.
    - Count all such subsegments where \( g \neq 0 \) after processing.

**Problem Specification**:

- **Constraints**:
    - \( 1 \leq t \leq 10^4 \)
    - \( 1 \leq n \leq 2 \times 10^5 \)
    - \( 1 \leq x \leq 10^9 \)
    - \( 1 \leq a_i \leq 10^9 \) for all \( i \) from \( 1 \) to \( n \).
    - The total sum of \( n \) across all test cases does not exceed \( 2 \times 10^5 \).

- **Additional Notes**:
    - A subsegment is defined by two indices \( l \) and \( r \) such that \( 1 \leq l \leq r \leq n \).
    - The goal is to efficiently compute the number of valid subsegments for each test case, adhering to the given constraints.
    - Optimal algorithms may be required to handle the large input size within reasonable time limits.
\\

\hrulefill
\\

\textbf{Deepseek-Coder-33B:} \\
"Problem Requirements": \\
1. "Input Format": 
Each test consists of multiple test cases. The first line contains an integer \( t \) (\( 1 \le t \le 10^4 \)) — the number of test cases. Then follows the description of the test cases.
- The first line of each test case contains two integers \( n \), \( x \) (\( 1 \leq n \leq 2 \cdot 10^5, 1 \le x \le 10^9 \)) — the number of mushrooms and the maximum toxicity level.
- The second line of each test case contains \( n \) numbers \( a_1, a_2, \ldots, a_n \) (\( 1 \leq a_i \leq 10^9 \)).

It is guaranteed that the sum of \( n \) over all test cases does not exceed \( 2 \cdot 10^5 \).

2. "Output Format": 
For each test case, output a single number — the number of subsegments such that the final value of \( g \) will not be zero.

3. "Functionality": 
Yaroslav is playing a computer game where he encounters \( n \) mushrooms with varying toxicity levels arranged in a row. The task is to find the number of subsegments such that the character's final toxicity level is not zero after consuming mushrooms within that subsegment. The character's toxicity level increases as mushrooms are consumed, but if it exceeds the maximum toxicity level, it resets to zero.

"Problem Specification": 
Yaroslav faces a challenge in determining the count of subsegments where the character's toxicity level remains non-zero. The input includes the number of test cases, the number of mushrooms, the maximum toxicity level, and the toxicity levels of each mushroom. The output should provide the count of such subsegments for each test case. The algorithm should consider the sequential consumption of mushrooms and the toxicity level constraints to calculate the desired outcome accurately.
\\
\blankspace
\\
\textbf{Teacher Programmer:} \\
You have understood the problem statement, including the requirements and specifications.

\end{promptbox}

\clearpage

\begin{promptbox}[Deepseek-Coder-33B generation at planning stage]{softblue}

\textbf{"Algorithm":} Sliding Window Technique. \\

\textbf{"Reasons for the algorithm choice":} \\
The Sliding Window Technique is suitable for this problem as it involves maintaining a window of variable size while iterating through the mushrooms. This method allows us to efficiently calculate the toxicity level and keep track of subsegments where the final toxicity level is non-zero. By adjusting the window boundaries based on toxicity levels and constraints, we can accurately count the desired subsegments within the given limits.

\textbf{"Algorithm Pseudocode":} \\
1. Initialize variables count = 0, g = 0, start = 0. \\
2. Iterate through the mushrooms from 0 to n-1 using a loop: \\
   - Increment g by the toxicity level of the current mushroom. \\
   - While g is greater than x, reduce g by the toxicity level of the mushroom at the start index and increment start. \\
   - If g is not equal to 0, update count by adding the size of the current window. \\
3. Output count as the number of subsegments where the final value of g is not zero. \\

\end{promptbox}

\begin{promptbox}[Teacher programmer feedback at planning stage]{softblue}

\textbf{Teacher Programmer :} \\
"Unsatisfied": "I think you have choosen the wrong algorithm. Please modify your answer accordingly."\\

**"Algorithm":** Two-Pointer Sliding Window

**"Reasons for the algorithm choice":**

1. **Efficiency**: The two-pointer sliding window technique operates in linear time \( O(n) \), which is optimal given the constraints (\( n \leq 2 \times 10^5 \) per test case and \( t \leq 10^4 \)).

2. **Simplicity**: This method is straightforward to implement and understand, reducing the likelihood of errors during coding.

3. **Applicability**: The problem essentially reduces to counting the number of subarrays with a sum \( \leq x \). The sliding window approach is well-suited for such range-based sum queries, especially when dealing with only positive integers (as toxicity levels \( a_i \) are \( \geq 1 \)).

4. **Optimal for Constraints**: Given that the sum of \( n \) across all test cases does not exceed \( 2 \times 10^5 \), an \( O(n) \) per test case algorithm ensures that the solution will run efficiently within the time limits.

\hrulefill
\\

\textbf{Deepseek-Coder-33B:} \\
\textbf{"Algorithm":} Two-Pointer Sliding Window \\

\textbf{"Reasons for the algorithm choice":} \\
The decision to utilize the sliding window algorithm was based on a thorough analysis of various factors. Firstly, the algorithm's linear time complexity of \(O(n)\) aligns well with the constraints of the problem, ensuring optimal efficiency given the input size limitations (\(n \leq 2 \times 10^5\) per test case and \(t \leq 10^4\)). Secondly, its straightforward implementation and intuitive nature simplify the coding process, reducing the chances of errors. Additionally, the problem's nature of counting subarrays with a sum less than or equal to a given threshold \(x\) makes the sliding window technique particularly well-suited, especially when dealing with non-negative integers (as indicated by the toxicity levels \(a_i \geq 1\)). Lastly, considering the cumulative sum of \(n\) across all test cases does not exceed \(2 \times 10^5\), an \(O(n)\) per test case approach ensures that the solution remains efficient and within the specified time limits.
\\

\textbf{"Algorithm Pseudocode":} 
\begin{verbatim}
function slidingWindow(arr, target):
    left = 0
    result = some initial value 
    for right from 0 to length(arr) - 1:
    
        windowValue = arr[right]  

        while some condition on the window (e.g., sum > target, size of window exceeds limit):

            windowValue += arr[left]
            left -= 1  # Move the left pointer to the right
       
        result = updateResult(result, windowValue)  
    return result
\end{verbatim}

\end{promptbox}

\clearpage

\begin{promptbox}[Teacher programmer feedback at planning stage]{softblue}

\textbf{Teacher Programmer Generation:} \\
"Unsatisfied": "I think the your pseudocode has some errors, please modify it accordingly"\\

\textbf{Pseudocode:} \\  

\begin{verbatim}
for each test case:
    # Step 1: Read inputs
    read n and x   # n is the size of the array, x is the sum constraint
    read array a of size n  # Array of size n

    # Step 2: Initialize variables
    left = 0  # Left pointer of the sliding window
    current_sum = 0  # The sum of elements in the current window
    total_subsegments = 0  # To store the count of subarrays with sum <= x

    # Step 3: Traverse the array using the right pointer
    for right from 0 to n-1:
        # Add the current element to the window
        current_sum += a[right]

        # Step 4: Shrink the window if the sum exceeds x
        while current_sum > x and left <= right:
            # Remove the leftmost element from the window
            current_sum -= a[left]
            # Move the left pointer to the right
            left += 1

        # Step 5: Count all valid subarrays ending at 'right'
        # All subarrays starting from 'left' to 'right' have sum <= x
        total_subsegments += (right - left + 1)

    # Step 6: Output the result for this test case
    output total_subsegments
\end{verbatim}

\hrulefill
\\

\textbf{Deepseek-Coder-33B:} \\
OK, here is my new answer: \\

\begin{verbatim}
for each test case:
    read n and x   # n is the size of the array, x is the sum constraint
    
    read array a of size n

    left = 0  # left pointer of the sliding window
    current_sum = 0  # the sum of elements in the current window
    total_subsegments = 0  # count of valid subarrays with sum <= x

    for right from 0 to n-1:
        # Add the current element to the current sum
        current_sum += a[right]

        while current_sum > x and left <= right:
            # Remove the leftmost element from the current sum
            current_sum -= a[left]
            # Move the left pointer to the right
            left += 1

        # All subarrays starting from left to right have sum <= x
        total_subsegments += (right - left + 1)

    output total_subsegments

\end{verbatim}

\hrulefill
\\
\textbf{Teacher Programmer:} \\
"Satisfied": You have chosen the correct algorithm and generated the corresponding accurate pseudocode."

\end{promptbox}

\clearpage

\begin{promptbox}[Deepseek-Coder-33B generation at coding stage]{softblue}

\begin{verbatim}
# Function to solve the problem for each test case
def solve_test_case(n, x, a):
    left = 0
    current_sum = 0
    total_subsegments = 0
    
    for right in range(n):
        current_sum += a[right]
        
        while current_sum > x and left <= right:
            current_sum -= a[left]
            left += 1
        
        total_subsegments += (right - left + 1)
    
    return total_subsegments

# Main function to read input and process test cases
def main():
    t = int(input())
    
    for _ in range(t):
        n, x = map(int, input().split())
        a = list(map(int, input().split()))
        
        result = solve_test_case(n, x, a)
        print(result)

# Run the main function
if __name__ == "__main__":
    main()
\end{verbatim}

\end{promptbox}

\clearpage

\begin{promptbox}[Deepseek-Coder-33B generation at coding stage]{softblue}

\textbf{Teacher Programmer:} \\

"Unsatisfied" : "The solution provided has some errors and misunderstandings that need to be addressed for it to correctly solve the problem. Let's break down the issues and provide suggestions for modifications."

Suggestions for the Solution

1. **Understand the reset condition**:
   - For each subsegment `[l, r]`, we need to track whether the cumulative sum exceeds `x`. If it does, we reset `g` to zero and stop counting subsegments starting from that index.
   - We need to count the number of valid subsegments where the cumulative sum never exceeds `x`.

2. **Sliding window for subsegments**:
   - Instead of just moving the left pointer whenever the sum exceeds `x`, we need to adjust the window and ensure that the sum never exceeds `x` at any point during the traversal of the subsegment.
   - Specifically, if the sum goes above `x`, the subsegment is invalid, and we should break out of the loop.

3. **Efficiently count valid subsegments**:
   - For each test case, we should compute all possible subsegments efficiently, taking into account the maximum possible sum `x`. For a subsegment to be valid, the cumulative sum should not exceed `x` at any time.

4. **Optimization with prefix sums**:
   - One approach would be to maintain a prefix sum and calculate the sum for each subsegment `[l, r]`. This allows for fast sum calculation without needing to sum over each range repeatedly.
   - The sliding window can be adjusted dynamically by checking whether the sum exceeds `x` as we move through the array.

General Steps for the New Solution:

1. **Iterate through each test case**.
2. For each test case:
   - Initialize `left` pointer and set `current\_sum` to 0.
   - Use the `right` pointer to iterate through the array.
   - For each `right`, add `a[right]` to `current\_sum`.
   - While `current\_sum` exceeds `x`, move the `left` pointer right to shrink the window.
   - For each valid window (where `current\_sum <= x`), count all subarrays that end at `right` and start from any index between `left` and `right`.
3. Print the result for each test case.

This approach ensures that the problem is solved in an efficient manner, using sliding window techniques, and counting the valid subsegments in a way that satisfies the problem's constraints.

Complexity Considerations:
- The sliding window approach ensures that both the `left` and `right` pointers move from 0 to `n-1`, making the solution efficient with a time complexity of `O(n)` for each test case.
- Given that the total sum of `n` across all test cases is constrained to \(\ 2 \times 10^5 \), the approach is efficient enough for the input limits.

\hrulefill

Deepseek-Coder-33B:
Here is my revision code.
\begin{verbatim}
def solve_test_case(n, x, a):
    left = 0
    current_sum = 0
    total_subsegments = 0

    # Use sliding window to count valid subsegments
    for right in range(n):
        current_sum += a[right]  # Expand the window by adding the current element
        # Shrink the window if the current_sum exceeds x
        while current_sum > x:
            current_sum -= a[left]
            left += 1
        # Count all valid subsegments ending at 'right'
        total_subsegments += (right - left + 1)
    return total_subsegments

def main():
    # Read the number of test cases
    t = int(input())
    results = []
    for _ in range(t):
        # Read n and x
        n, x = map(int, input().split())
        # Read array a
        a = list(map(int, input().split()))
        
        # Solve for the current test case and store the result
        result = solve_test_case(n, x, a)
        results.append(result)
    # Print all results
    print("\n".join(map(str, results)))
if __name__ == "__main__":
    main()
\end{verbatim}

"Satisfied:" "You have chosen the correct solution."

\end{promptbox}

\clearpage

\begin{promptbox}[Deepseek-Coder-33B generation at debugging stage]{softblue}

"Based on the compiler output, I modify my code." \\
"Bug Identification"

Reset condition handling: If the cumulative sum exceeds x, not only should the left pointer adjust, but we also need to ensure the sum resets properly so invalid subsegments are not included. 

Valid subsegment count: When toxicity resets (g becomes zero), all subsegments involving the invalid segment should be discarded. 

"problem Resolution:"

\begin{verbatim}
# Function to solve the problem for each test case
def solve_test_case(n, x, a):
    left = 0
    current_sum = 0
    total_subsegments = 0

    for right in range(n):
        current_sum += a[right]

        # If toxicity exceeds x, reset and adjust the left pointer
        while current_sum > x and left <= right:
            current_sum -= a[left]
            left += 1
        
        # Only count valid subsegments where the current_sum never exceeds x
        total_subsegments += (right - left + 1)
    
    return total_subsegments

def main():
    # Read number of test cases
    t = int(input())
    results = []

    for _ in range(t):
        n, x = map(int, input().split())
        a = list(map(int, input().split()))
        # Solve and collect results
        results.append(solve_test_case(n, x, a))
    
    # Output all results
    print("\n".join(map(str, results)))

if __name__ == "__main__":
    main()

\end{verbatim}

\end{promptbox}

\clearpage

\begin{promptbox}[Teacher programmer feedback at debugging stage]{softblue}

\textbf{Teacher Programmer:}
**"unsatisfied":** "There are some syntactic(semantic) errors in your code. Here are my suggestions for modifications:

- **Syntactic errors**:

  1. **Function related errors**: Your code has function-related errors in the `solve\_test\_case` function, where it incorrectly calculates the total number of subsegments. This results in a wrong value being returned when reset conditions are not handled correctly.
  
  2. **Operation errors**: Your code uses the `+` operation in a context where the cumulative sum logic does not account for invalid subsegments, leading to an overcount of subsegments in certain cases.

- **Semantic errors**:

  1. **Control flow errors**: Your code has control flow errors in the while-loop condition within `solve\_test\_case`, which fails to efficiently handle the reset condition. As a result, the toxicity reset logic does not properly skip over invalid ranges.
  
  2. **Logical direction errors**: Your code counts all subsegments ending at `right` even after a reset occurs, which is incorrect. Valid subsegments should only include those where the cumulative sum is within the limit.

Please rewrite your code accordingly."

\blankspace

\textbf{Deepseek-Coder-33B:}

\begin{verbatim}
def solve(n, x, a):
    nextlex = [0] * n
    i = 0
    j = 1
    while i < n - 1:
        if j == n - 1:
            nextlex[i] = j
            i += 1
        elif j == i:
            j += 1
        else:
            if a[j] <= x:
                nextlex[i] = j
                i += 1
            else:
                j += 1
    pointto = {}
    count = [0] * n
    i, j, sumi = 0, 0, 0
    while i < n:
        if j == n:
            count[i] = j - i
            sumi -= a[i]
            i += 1
        elif sumi + a[j] <= x:
            sumi += a[j]
            j += 1
        else:
            count[i] = j - i
            if j < n - 1:
                pointto[i] = nextlex[j]
            sumi -= a[i]
            i += 1
    for i in range(n - 1, -1, -1):
        if i in pointto:
            count[i] += count[pointto[i]]
    return sum(count)
def main():
    t = int(input())
    results = []
    for _ in range(t):
        n, x = map(int, input().split())
        a = list(map(int, input().split()))
        results.append(solve(n, x, a))
    print("\n".join(map(str, results)))
if __name__ == "__main__":
    main()
\end{verbatim}

\blankspace

\textbf{Teache Programmer:}
"Satisfied": "You’ve fixed all the bugs in your code."

\end{promptbox}

\clearpage

\clearpage

\section{Error Analysis}
\label{errrrr}

\subsection{Error Classification}
\begin{table}[H]
\small
\centering
\renewcommand{\arraystretch}{1.5} 
\resizebox{\textwidth}{!}{%
\renewcommand{\arraystretch}{1.5}
\begin{tabular}{p{0.2\textwidth}|p{0.2\textwidth}|p{0.5\textwidth}}
\toprule
\textbf{Error Category} & \textbf{Error Type} & \textbf{Error Explanation} \\ 
\midrule
\multirow{12}{*}{\textbf{Syntactic Errors}}
& \multirow{2}{*}{Function Related Errors} 
    & 1. Return Error: Returns a wrong value in an unexpected format. \\
& & 2. Function Call Error: Incorrect function name, wrong arguments, or incorrect method call target. \\
\cmidrule{2-3}
& Operation Errors 
    & Operation is applied to an inappropriate data type. \\ 
\cmidrule{2-3}
& \multirow{4}{*}{Structure Errors} 
    & 1. Code Block Error: Incorrectly generated or omitted statements, leading to task failure. \\
& & 2. Punctuation Error: Errors in punctuation, such as missing semicolons, commas, or brackets. \\
\cmidrule{2-3}
& \multirow{2}{*}{Declaration Errors} 
    & Declaration Error: Incorrect or duplicate declaration of variables or functions. \\
\cmidrule{2-3}
& \multirow{2}{*}{Import Errors} 
    & Library/Module Import Error: Failure to import external libraries or nonexistent library/module. \\ 
\midrule

\multirow{12}{*}{\textbf{Semantic Errors}}   
& \multirow{2}{*}{Control Flow Errors}  
    & 1. Condition Error: Logical error in a conditional statement causing unexpected execution. (if, else) \\
& & 2. Loop Error: Error leading to infinite loops or incorrect loop exits. (while, for) \\
\cmidrule{2-3}
& \multirow{2}{*}{Reference Errors} 
    & 1. Wrong Function: Calling the wrong function. \\
& & 2. Wrong Variable: Calling the wrong variable. \\
\cmidrule{2-3}
& \multirow{2}{*}{Calculation Errors} 
    & 1. Incorrect Arithmetic Operation. \\
& & 2. Incorrect Comparison Operation. \\ 
\cmidrule{2-3}
& \multirow{2}{*}{Incomplete Errors} 
    & Missing essential logical steps: The code generation is incomplete. \\
    
\cmidrule{2-3}
& \multirow{2}{*}{Logical Direction Error} 
    & The code significantly deviates from intended logic and expected outcomes. \\ 

\cmidrule{2-3}
& Suboptimal Errors 
    & Suboptimal solutions lead to exceeding time or memory limits.  \\
\bottomrule
\end{tabular}}
\caption{Taxonomy of syntactic/semantic characteristics of code errors made by LLMs}
\label{error_taxonomy}
\end{table}

\clearpage

\subsection{Real Human Debug}

\begin{promptbox}[Guidelines for human programmers to identify code errors and provide suggestions for problem resolution.]{lightgreen}
\label{guidelines for human}
You need to help identify code errors and provide suggestions based on the problem statement \{PROBLEM\_STATEMENT\}, input format \{INPUT\_FORMAT\}, output format \{OUTPUT\_FORMAT\}, test case example \{TEST\_CASE\_EXAMPLE\}, the python code \{CODE\} LLM programmer generated, the compiler output \{COMPILER\_OUTPUT\}, all the test cases {TEST\_CASE\}. 
\\
\\
Your feedback needs to focus on the following content:  \\

**Code Syntactic Correctness**: You should concentrate on errors such as those related to functions, operations, structure, declarations, and external dependencies. Here are the detailed classification: \\

1. Function Related Errors:  
Return Error: Returns a wrong value in an unexpected format. 
Function Call Error: Incorrect function name, wrong arguments, or incorrect method call target. \\

2. Operation Errors:
Operation is applied to an inappropriate data type.\\

3. Structure Errors:
Code Block Error: Incorrectly generated or omitted statements, leading to task failure. Punctuation Error: Errors in punctuation, such as missing semicolons, commas, or brackets. \\ 

4. Declaration Errors:
Declaration Error: Incorrect or duplicate declaration of variables or functions. \\

5. Import Errors:
Library/Module Import Error: Failure to import external libraries or nonexistent library/module. \\

**Code Semantic Correctness**: You should concentrate on issues related to control flow, references, calculations, incompleteness, logical direction and suboptimal errors. \\

1. Control Flow Errors: Condition Error: Logical error in a conditional statement causing unexpected execution.  Loop Error: Error leading to infinite loops or incorrect loop exits. (while, for).  \\
 
2. Wrong Function: Calling the wrong function. Wrong Variable: Calling the wrong variable. \\

3. Calculation Errors: Incorrect Arithmetic Operation. Incorrect Comparison Operation. \\

4. Incomplete Errors:  Missing essential logical steps: The code generation is incomplete.\\

5. Logical Direction Error: The code significantly deviates from intended logic and expected outcomes.\\

6. Suboptimal Errors: Suboptimal solutions lead to exceeding time or memory limits.\\

Please provide your feedback in the following format.\\

"Code Error Identifications:" \\
"Syntactic Errors:" ...\\
"Semantic Errors:" ...\\

"Resolution Suggestions:" ... \\

}\end{promptbox}

\clearpage

\begin{promptbox}[AtCoder Regular Contest 183 Problem C]{lightgreen}

\textbf{GPT-4-Turbo generated code: } \\
\begin{verbatim}
MOD = 998244353
def solve(N, M, conditions):
    dp = [[0] * (N + 1) for _ in range(N + 1)]
    dp[0][0] = 1

    for L, R, X in conditions:
        for i in range(N, 0, -1):
            for j in range(N, -1, -1):
                if j < L - 1 or j > R:
                    dp[i][j] = (dp[i][j] + dp[i-1][j]) % MOD
                elif j < X - 1 or j > R:
                    dp[i][j] = (dp[i][j] + dp[i-1][j-1]) % MOD
                elif X - 1 <= j <= R:
                    dp[i][j] = (dp[i][j] + dp[i-1][R]) % MOD

    return sum(dp[N]) % MOD
N, M = map(int, input().split())
conditions = []
for _ in range(M):
    L, R, X = map(int, input().split())
    conditions.append((L - 1, R - 1, X - 1))
result = solve(N, M, conditions)
print(result)
\end{verbatim}

\textbf{Automatic Debug:} \\

\textbf{Error Type: Logical Direction Errors} \\
\textbf{Resolution Suggestions:} \\
Off-by-One Error: The input uses 1-based indexing, but the code uses 0-based indexing. The conditions (L, R, X) should be adjusted properly.\\
Incorrect Dynamic Programming Transitions: The transitions in the dp table are not correctly defined based on the ranges (L, R, X), leading to incorrect updates.\\
Inefficient Looping: The dp table is updated from the wrong direction, which may overwrite needed values during the same iteration.\\
Adjust Indexing: Convert the input conditions (L, R, X) to 0-based indexing by subtracting 1 when reading input. \\
Correct Transitions in dp Table:
Update dp[i][j] based on the range conditions, correctly handling:
Values outside [L, R] inherit directly from dp[i-1][j].
Values within [L, X-1] inherit from dp[i-1][j-1].
Values within [X, R] inherit from dp[i-1][R]. \\
Use Backward Looping:
Process the dp table backwards (from N to 1 and j from N to 0) to avoid overwriting values that are still needed in the same iteration. \\

\textbf{GPT-4-Turbo Correction:}
\begin{verbatim}
MOD = 998244353
def solve(N, M, conditions):
    dp = [[0] * (N + 1) for _ in range(N + 1)]
    dp[0][0] = 1  # Base case

    for L, R, X in conditions:
        for i in range(N, 0, -1):
            for j in range(N, -1, -1):
                if j < L or j > R:
                    dp[i][j] = (dp[i][j] + dp[i-1][j]) % MOD
                elif L <= j < X:
                    dp[i][j] = (dp[i][j] + dp[i-1][j-1]) % MOD
                elif X <= j <= R:
                    dp[i][j] = (dp[i][j] + dp[i-1][R]) % MOD

    return sum(dp[N]) % MOD
# Reading input
N, M = map(int, input().split())
conditions = [tuple(map(lambda x: int(x)-1, input().split())) for _ in range(M)]
# Output the result
print(solve(N, M, conditions))
\end{verbatim}

\end{promptbox}

\clearpage

\begin{promptbox}[AtCoder Regular Contest 183 Problem C]{lightgreen}
\textbf{Human Debug} \\
\textbf{Semantic Errors: } \\
\textbf{Control Flow Errors:} \\

Your code still remain some control flow errors. \\
for length from 1 to n: Traverse the subsequence lengths, from 1 to n.\\
for i from 1 to n - length + 1: Traverse the starting positions i of the subsequences, ensuring the subsequence length does not exceed n. \\
j = i + length - 1: Calculate the ending position j of the subsequence based on the starting position i and the subsequence length.\\
for k from i to j: Traverse all possible split points k, dividing the subsequence [i, j] into two subranges: [i, k-1] and [k+1, j].\\
if g[i][j][k] is true then continue: If there is a constraint at position k within the range [i, j], skip that split point.\\
dp[i][j] = dp[i][j] + dp[i][k - 1] * dp[k + 1][j] * C[j - i][k - i]: Calculate the number of valid ways for the current subsequence [i, j], using the valid counts of the left subrange [i, k-1] and the right subrange [k+1, j], along with the combination count C[j - i][k - i].\\
dp[i][j] = dp[i][j] \% MOD: Take the result modulo MOD to prevent overflow.\\

\textbf{GPT-4-Turbo Correction}
\begin{verbatim}
MOD = 998244353
N = 510
n, m = map(int, input().split())
dp = [[0] * N for _ in range(N)]
for i in range(1, n + 2):
    dp[i][i - 1] = 1  # Initialize dp[i][i-1] as 1 (empty subsequence)
for length in range(1, n + 1):
    for i in range(1, n - length + 2):
        j = i + length - 1
        for k in range(i, j + 1):
            if g[i][j][k]:
                continue
            dp[i][j] = (dp[i][j] + dp[i][k - 1] * dp[k + 1][j] * C[j - i][k - i]) % MOD
print(dp[1][n])

\end{verbatim}

\end{promptbox}

\clearpage

\begin{promptbox}[AtCoder Regular Contest 183 Problem C]{lightgreen}
\textbf{Human Debug}  \\
\textbf{Incomplete Errors} \\
Your code has some incomplete errors, you need to complete the array g iteration. \\
Iterate over subinterval lengths: It loops through all possible subinterval lengths, starting from 2 up to n. The length represents the current length of the subinterval. \\
Iterate over starting positions: For each subinterval length, it loops through all valid starting positions i such that the subinterval does not exceed the array bounds. The loop runs from 1 to n - length + 1. \\
Calculate ending position: The ending position j of the current subinterval is calculated using the formula j = i + length - 1. \\
Iterate over possible split points: For each subinterval [i, j], it loops through all possible positions k (from i to j) within the subinterval. \\
Update constraint information: For each position k, it updates the value of g[i][j][k]. This represents whether position k in the subinterval [i, j] is affected by constraints. Specifically, the value is updated by checking if the position k is constrained in either the subinterval [i+1, j] or [i, j-1]. If either of these neighboring subintervals has a constraint at position k, the current subinterval [i, j] will also have a constraint at position k. \\

\textbf{GPT-4-Turbo Correction} \\
\begin{verbatim}

MOD = 998244353
N = 510

n, m = map(int, input().split())

C = [[0] * N for _ in range(N)]
for i in range(N):
    C[i][0] = C[i][i] = 1
    for j in range(1, i):
        C[i][j] = (C[i - 1][j] + C[i - 1][j - 1]) % MOD

g = [[[False] * N for _ in range(N)] for _ in range(N)]
for _ in range(m):
    l, r, x = map(int, input().split())
    g[l][r][x] = True

for length in range(2, n + 1):
    for i in range(1, n - length + 2):
        j = i + length - 1
        for k in range(i, j + 1):
            g[i][j][k] = g[i][j][k] or g[i + 1][j][k] or g[i][j - 1][k]

dp = [[0] * N for _ in range(N)]
for i in range(1, n + 2):
    dp[i][i - 1] = 1

for length in range(1, n + 1):
    for i in range(1, n - length + 2):
        j = i + length - 1
        for k in range(i, j + 1):
            if g[i][j][k]:
                continue
            dp[i][j] = (dp[i][j] + dp[i][k - 1] * dp[k + 1][j] * C[j - i][k - i]) % MOD
print(dp[1][n])

\end{verbatim}

\end{promptbox}

\clearpage

\subsection{Code Bug Annotation Example}
\label{bug_example}

\begin{promptbox}[AtCoder Regular Contest 183 Problem C]{lightgreen}

\textbf{GPT-4-Turbo generated code: } \\
\begin{verbatim}
MOD = 998244353

def solve(N, M, conditions):
    dp = [[0] * (N + 1) for _ in range(N + 1)]
    dp[0][0] = 1

    for L, R, X in conditions:
        for i in range(N, 0, -1):
            for j in range(N, -1, -1):
                if j < L - 1 or j > R:
                    dp[i][j] = (dp[i][j] + dp[i-1][j]) % MOD
                elif j < X - 1 or j > R:
                    dp[i][j] = (dp[i][j] + dp[i-1][j-1]) % MOD
                elif X - 1 <= j <= R:
                    dp[i][j] = (dp[i][j] + dp[i-1][R]) % MOD

    return sum(dp[N]) % MOD

N, M = map(int, input().split())
conditions = []
for _ in range(M):
    L, R, X = map(int, input().split())
    conditions.append((L - 1, R - 1, X - 1))

result = solve(N, M, conditions)
print(result)
\end{verbatim}

\end{promptbox}

\clearpage

\begin{promptbox}[AtCoder Regular Contest 183 Problem C]{lightgreen}

\textbf{Human Programmer Annotation:} \\
\textbf{Semantic erros:} \\
\textbf{Control Flow Errors and Incomplete Errors:} \\
This code has control flow errors and incomplete errors in the following section:

\begin{verbatim}[language=Python]
if j < L - 1 or j > R:
    dp[i][j] = (dp[i][j] + dp[i-1][j]) % MOD
elif j < X - 1 or j > R:
    dp[i][j] = (dp[i][j] + dp[i-1][j-1]) % MOD
elif X - 1 <= j <= R:
    dp[i][j] = (dp[i][j] + dp[i-1][R]) % MOD
\end{verbatim}

\textbf{Correction Suggestions:}
\begin{verbatim}[language=Python]
# Loop through each possible subsequence length w
for w in range(2, n + 1):
    # Initialize an array y to keep track of constraints for length w
    y = [0] * n

    # Preprocess constraints that end before the current range
    for r in range(w - 1):
        for ll, x in pr[r]:
            # Increment the constraint count for y[x]
            y[x] += 1

    # Iterate over all valid starting indices l for a subsequence of length w
    for l in range(n - w + 1):
        r = l + w - 1  # Calculate the end index of the current subsequence

        # Consider constraints ending at r
        for ll, x in pr[r]:
            if l <= ll:
                y[x] += 1

        # Calculate number of valid sequences for the current subsequence l to r
        for x in range(l, r + 1):
            if y[x] == 0:  # Check if there are no active constraints at x
                if x == l:
                    q[l][r] += q[x + 1][r]
                elif x == r:
                    q[l][r] += q[l][x - 1]
                else:
                    # Calculate the number of ways to split the sequence at x
                    q[l][r] += (q[l][x - 1] * q[x + 1][r] * fa[r - l] * fb[x - l] * fb[r - x]) % M

                # Take modulo to prevent overflow
                q[l][r] %= M

        # Adjust constraints as we move the starting index l
        for rr, x in pl[l]:
            if rr <= r:
                y[x] -= 1
\end{verbatim}

\end{promptbox}

\clearpage

\begin{promptbox}[AtCoder Regular Contest 180 Problem D]{lightgreen}

\textbf{GPT-4-Turbo Generated Code:} \\

\begin{verbatim}[language=Python]
def min_partition_sum(A, L, R):
    if R - L + 1 < 3:
        return float('inf')

    max_left = max(A[L:L + 2])
    max_right = max(A[R-2:R+1]) 
    max_middle = float('-inf')
    min_sum = float('inf')

    for i in range(L + 2, R - 1):  
        max_middle = max(max_middle, A[i])
        current_sum = max_left + max_middle + max_right
        min_sum = min(min_sum, current_sum)

        if A[i - 1] == max_left:
            max_left = max(A[L + 2:i])
        else:
            max_left = max(max_left, A[L + 2:i - 1])

    max_middle = max(max_middle, A[R - 1])  
    current_sum = max_left + max_middle + max_right
    min_sum = min(min_sum, current_sum)

    max_middle = max(max(A[L + 1:R - 1]), A[R - 1])
    current_sum = max_left + max_middle + max(A[R - 2:R])
    min_sum = min(min_sum, current_sum)

    return min_sum

def main():
    N, Q = map(int, input().split())  
    A = list(map(int, input().split()))  

    results = []

    for _ in range(Q): 
        L, R = map(int, input().split())
        results.append(min_partition_sum(A, L - 1, R - 1))  

    for result in results:
        print(result)  

if __name__ == "__main__":
    main()

\end{verbatim}

\end{promptbox}

\clearpage

\begin{promptbox}[AtCoder Regular Contest 180 Problem D]{lightgreen}
\textbf{Human Annotation: }\\
\textbf{Syntactic Errors: }\\
\textbf{Operation Errors:} \\
This code has control flow errors in the following section:
max\_left = max(max\_left, A[L + 2:i - 1]), '>' not supported between instances of 'list' and 'int' . \\

\textbf{Correction Suggestions:}
\begin{verbatim}[language=Python]
# Find the minimum sum of max values by trying different partitions
    for i in range(L + 2, R):  # Ensure we don't go out of bounds
        max_middle = max(max_middle, A[i])
        current_sum = max_left + max_middle + max_right
        min_sum = min(min_sum, current_sum)

        # Update max_left based on the current position
        if A[i - 1] == max_left:
            # Only recalculate if max_left needs updating
            if i - 2 >= L + 2:  # Ensure we don't go out of bounds
                max_left = max(A[L + 2:i - 1])
        else:
            max_left = max(max_left, A[L + 2:i - 1])
\end{verbatim}

Ensure max\_left is Always an Integer: \\
When updating the position of max\_left, ensure that the parameters used in the max function are valid numeric values and avoid passing a list to the max function.\\
Boundary Condition Checks: 

When calculating max\_left, add boundary condition checks to ensure that accessing the list does not go out of bounds.
Clarification of Logic:

In the logic for updating max\_left, clarify the conditions and the updating method to ensure that max\_left consistently receives the correct value.

\end{promptbox}
\clearpage

\section{Prompt}
\label{sec:protocol_prompt}

\begin{promptbox}[System prompt for LLMs at comprehension stage]{lightgreen}
\label{tab:comprehension_prompt}
You are a competitive programming programmer who gives the problem comprehension based on the problem statement \{PROBLEM\_STATEMENT\}, input format \{INPUT\_FORMAT\}, output format \{OUTPUT\_FORMAT\}, test case example \{TEST\_CASE\_EXAMPLE\}. 
\\
\\
Your comprehension needs to focus on the following content:  \\

1. **Problem Requirements**: Problem Requirements refer to the necessary input and output formats that must be followed, as well as the specific functionalities that the program is required to implement.

2. **Problem Specifications**: Problem Specifications refer to the specific conditions and constraints that must be met for a solution to a problem to be considered valid and effective. \\

Your output must follow the format below. \\

"Problem Requirements": ... \\
 1. "Input:" ... \\
 2. "Output" ... \\
 3. "Functionality..." \\
 \\
"Problem Specification": ...\\

\end{promptbox}

\begin{promptbox}[System prompt for LLMs at planning stage]{lightgreen}
\label{tab:planning_prompt}
You are a competitive programming programmer who chooses algorithm to solve this problem based on the problem statement \{PROBLEM\_STATEMENT\}, input format \{INPUT\_FORMAT\}, output format \{OUTPUT\_FORMAT\}, test case example \{TEST\_CASE\_EXAMPLE\}, and problem requirements \{PROBLEM\_REQUIREMENTS\}, problem specification \{PROBLEM\_SPECIFICATION\} you generated in comprehension stage.
\\
\\
The algorithm you choose needs to focus on the following content:  \\

1. **Algorithm Correctness**: Algorithm you choose should be effective for this problem, your goal is to pass all the test cases.

Your output must follow the format below. \\

"Algorithm": ... \\
"Reasons for the algorithm choice": ...\\
"Algorithm psedocode:"... \\

\end{promptbox}

\begin{promptbox}[System prompt for LLMs at coding stage]{lightgreen}
\label{tab:coding_prompt}
You are a competitive programming programmer who generates python code to solve this problem based on the problem statement \{PROBLEM\_STATEMENT\}, input format \{INPUT\_FORMAT\}, output format \{OUTPUT\_FORMAT\}, test case example \{TEST\_CASE\_EXAMPLE\}, problem requirements \{PROBLEM\_REQUIREMENTS\}, problem specification \{PROBLEM\_SPECIFICATION\} you generated in comprehension stage, algorithm \{ALGORITHM\} you choose in the planning stage. 
\\
\\
note you must not return anything except for the python code.

\end{promptbox}

\clearpage

\begin{promptbox}[System prompt for LLMs at debugging stage]{lightgreen}
\label{tab:debugging_prompt}
You are a competitive programming programmer who corrects python code errors based on the problem statement \{PROBLEM\_STATEMENT\}, input format \{INPUT\_FORMAT\}, output format \{OUTPUT\_FORMAT\}, test case example \{TEST\_CASE\_EXAMPLE\}, problem requirements \{PROBLEM\_REQUIREMENTS\}, problem specification \{PROBLEM\_SPECIFICATION\} you generated in comprehension stage, algorithm \{ALGORITHM\} you chose in the planning stage, python code \{CODE\} you generated in coding stage and the compiler output \{COMPILER\_OUTPUT\}. 
\\
\\
You need to follow these steps to debug. \\

1. **Bug Identification**: Bug identification involves locating and diagnosing errors within your Python code. \\
2. **Problem Resolution**: Problem Resolution refers to making modifications to the code based on the identified bugs.
\\

Your output must follow the format below. You must not return anything except for the python code for problem resolution.\\

"Bug Identification": ... \\
"Problem Resolution": ...\\

\end{promptbox}

\clearpage

\begin{promptbox}[System prompt for student programmer gives feedback at comprehension stage]{lightgreen}
\label{tab:uncertain_comprehension}
You are a competitive programming programmer who gives feedback to clarify the problem statement for the LLM programmer based on the problem statement \{PROBLEM\_STATEMENT\}, input format \{INPUT\_FORMAT\}, output format \{OUTPUT\_FORMAT\}, test case example \{TEST\_CASE\_EXAMPLE\} , the problem requirements \{PROBLEM\_REQUIREMENTS\} and problem specifications \{PROBLEM\_SPECIFICATION\} the LLM programmer generated.
\\
\\
Your feedback needs to focus on the following content:  \\

1. **Problem Requirements**: Problem Requirements refer to the necessary input and output formats that must be followed, as well as the specific functionalities that the program is required to implement.

2. **Problem Specifications**: Problem Specifications refer to the specific conditions and constraints that must be met for a solution to a problem to be considered valid and effective. \\

Your feedback must follow the one of the format below.

"satisfied": "You have understood the problem statement, including the requirements and specifications.", \\

"unsatisfied": "You have not understood the problem statement, missing the requirements and specifications. Let me clarify the problem statement for you ..., please modify your answer"  \\

\end{promptbox}

\begin{promptbox}[System prompt for student programmer gives feedback at planning stage]{lightgreen}
\label{tab:uncertain_planning}
You are a competitive programming programmer who gives feedback to provide effective algorithms for the LLM programmer based on the problem statement \{PROBLEM\_STATEMENT\}, input format \{INPUT\_FORMAT\}, output format \{OUTPUT\_FORMAT\}, test case example \{TEST\_CASE\_EXAMPLE\}, the algorithm \{ALGORITHM\} LLM chose ,the reason why LLM choose this algorithm \{REASONS\}, the algorithm pseducode \{PSEDOCODE\} LLM programmer generated. 
\\
\\
Your feedback needs to focus on the following content:  \\

1. **Algorithm Correctness**: Algorithm Correctness refers to whether the LLM programmer has chosen an appropriate algorithm for the current problem, including whether the rationale behind the LLM's choice of algorithm is reasonable.

2. **Pseudocode Correctness**: Pseudocode Correctness refers to whether the LLM programmer can generate accurate pseudocode code for the chosen algorithm, testing whether the LLM truly understands the selected algorithm.

Your feedback must follow the one of the format below. 
Note: You can provide the relevant algorithm and details to the LLM. You cannot directly provide pseudocode; you can only point out the incorrect parts of the pseudocode.

"satisfied": "You have chosen the correct algorithm and generated the corresponding accurate pseudocode.", \\

"unsatisfied": "I think you have choosen the wrong algorithm... or the reason why you choose this algorithm is not correct... or the pseudocode you generated has some errors..., please modify your answer"  \\

\end{promptbox}

\begin{promptbox}[System prompt for student programmer gives feedback at coding stage]{lightgreen}
\label{tab:uncertain_coding}
You are a competitive programming programmer who gives feedback to provide problem solutions for the LLM programmer based on the problem statement \{PROBLEM\_STATEMENT\}, input format \{INPUT\_FORMAT\}, output format \{OUTPUT\_FORMAT\}, test case example \{TEST\_CASE\_EXAMPLE\}, the python code \{CODE\} LLM programmer generated.
\\
\\
Your feedback needs to focus on the following content:  \\

1. **Solution Correctness**: Evaluate the correctness of the code logic generated by the LLM programmer to determine if it follows the correct approach.

Your feedback must follow the one of the format below. 
Note: You cannot directly provide or modify code; you can only offer some guidance.

"satisfied": "You have chosen the correct solution", \\

"unsatisfied": "There are some errors in your solution. Here are my suggestions for modifications... Please rewrite your code accordingly."  \\

\end{promptbox}

\clearpage

\begin{promptbox}[System prompt for student programmer gives feedback at debug stage]{lightgreen}
\label{tab:uncertain_debug}
You are a competitive programming programmer who gives feedback to help debug the code errors based on the problem statement \{PROBLEM\_STATEMENT\}, input format \{INPUT\_FORMAT\}, output format \{OUTPUT\_FORMAT\}, test case example \{TEST\_CASE\_EXAMPLE\}, the python code \{CODE\} LLM programmer generated, the compiler output \{COMPILER\_OUTPUT\}
\\
\\
Your feedback needs to focus on the following content:  \\

**Code Syntactic Correctness**: You should concentrate on errors such as those related to functions, operations, structure, declarations, and external dependencies. Here are the detailed classification: \\

1. Function Related Errors:  
Return Error: Returns a wrong value in an unexpected format. 
Function Call Error: Incorrect function name, wrong arguments, or incorrect method call target. \\

2. Operation Errors:
Operation is applied to an inappropriate data type.\\

3. Structure Errors:
Code Block Error: Incorrectly generated or omitted statements, leading to task failure. Punctuation Error: Errors in punctuation, such as missing semicolons, commas, or brackets. \\ 

4. Declaration Errors:
Declaration Error: Incorrect or duplicate declaration of variables or functions. \\

5. Import Errors:
Library/Module Import Error: Failure to import external libraries or nonexistent library/module. \\

**Code Semantic Correctness**: You should concentrate on issues related to control flow, references, calculations, incompleteness, logical direction and suboptimal errors. \\

1. Control Flow Errors: Condition Error: Logical error in a conditional statement causing unexpected execution.  Loop Error: Error leading to infinite loops or incorrect loop exits. (while, for).  \\
 
2. Wrong Function: Calling the wrong function. Wrong Variable: Calling the wrong variable. \\

3. Calculation Errors: Incorrect Arithmetic Operation. Incorrect Comparison Operation. \\

4. Incomplete Errors:  Missing essential logical steps: The code generation is incomplete.\\

5. Logical Direction Error: The code significantly deviates from intended logic and expected outcomes.\\

6. Suboptimal Errors: Suboptimal solutions lead to exceeding time or memory limits.\\

Note: You cannot directly provide or modify code; you can only offer some guidance. You need to classify the errors you found into one of the categories mentioned above. Your feedback must follow the one of the format below. 

"satisfied": "You've fixed all the bugs in your code. ", \\

"unsatisfied": "There are some syntactic(semantic) errors in your code. Here are my suggestions for modifications... 

syntactic errors: 1. function related errors: your code has function related errors in ..., which return a wrong value ...
                  2. operation errors: your code has function related errors in ... 
                  ..., which "+" operation can not add string type and int type.

semantic errors: 1. control flow errors: your code has control flow errors in..., which leads to incorrect if condition...

Please rewrite your code accordingly."  \\

\end{promptbox}

\clearpage

\begin{promptbox}[System prompt for teacher programmer gives feedback at comprehension stage]{lightgreen}
\label{tab:oracle_comprehension}
You are a competitive programming programmer who gives feedback to clarify the problem statement for the LLM programmer based on the problem statement \{PROBLEM\_STATEMENT\}, input format \{INPUT\_FORMAT\}, output format \{OUTPUT\_FORMAT\}, test case example \{TEST\_CASE\_EXAMPLE\} , the problem requirements \{PROBLEM\_REQUIREMENTS\} and problem specifications \{PROBLEM\_SPECIFICATION\} the LLM programmer generated, the correct requirements and specifications in \ourdata {Annotated\_Statement}.
\\
\\
Your feedback needs to focus on the following content:  \\

1. **Problem Requirements**: Problem Requirements refer to the necessary input and output formats that must be followed, as well as the specific functionalities that the program is required to implement.

2. **Problem Specifications**: Problem Specifications refer to the specific conditions and constraints that must be met for a solution to a problem to be considered valid and effective. \\

Your feedback must follow the one of the format below.

"satisfied": "You have understood the problem statement, including the requirements and specifications.", \\

"unsatisfied": "You have not understood the problem statement, missing the requirements and specifications. Let me clarify the problem statement for you ..."  \\

\end{promptbox}

\begin{promptbox}[System prompt for teacher programmer gives feedback at planning stage]{lightgreen}
\label{tab:oracle_planning}
You are a competitive programming programmer who gives feedback to provide effective algorithms for the LLM programmer based on the problem statement \{PROBLEM\_STATEMENT\}, input format \{INPUT\_FORMAT\}, output format \{OUTPUT\_FORMAT\}, test case example \{TEST\_CASE\_EXAMPLE\}, the algorithm \{ALGORITHM\} LLM chose ,the reason why LLM choose this algorithm \{REASONS\}, the algorithm pseducode \{PSEDOCODE\} LLM programmer generated, the correct algorithm \{ANNOTATED\_ALGORITHM\} and psedocode \{ANNOTATED\_PSEDOCODE\} in \ourdata. 
\\
\\
Your feedback needs to focus on the following content:  \\

1. **Algorithm Correctness**: Algorithm Correctness refers to whether the LLM programmer has chosen an appropriate algorithm for the current problem, including whether the rationale behind the LLM's choice of algorithm is reasonable.

2. **Pseudocode Correctness**: Pseudocode Correctness refers to whether the LLM programmer can generate accurate pseudocode code for the chosen algorithm, testing whether the LLM truly understands the selected algorithm.

Your feedback must follow the one of the format below. 
Note: You can provide the relevant algorithm and details to the LLM. You cannot directly provide pseudocode; you can only point out the incorrect parts of the pseudocode.

"satisfied": "You have chosen the correct algorithm and generated the corresponding accurate pseudocode.", \\

"unsatisfied": "I think you have choosen the wrong algorithm... or the reason why you choose this algorithm is not correct... or the pseudocode you generated has some errors..."  \\

\end{promptbox}

\begin{promptbox}[System prompt for teacher programmer gives feedback at coding stage]{lightgreen}
\label{tab:oracle_coding}
You are a competitive programming programmer who gives feedback to provide problem solutions for the LLM programmer based on the problem statement \{PROBLEM\_STATEMENT\}, input format \{INPUT\_FORMAT\}, output format \{OUTPUT\_FORMAT\}, test case example \{TEST\_CASE\_EXAMPLE\}, the python code \{CODE\} LLM programmer generated, the ground\_truth solutions \{GROUND\_TRUTH\_SOLUTION\} in \ourdata .
\\
\\
Your feedback needs to focus on the following content:  \\

1. **Solution Correctness**: Evaluate the correctness of the code logic generated by the LLM programmer to determine if it follows the correct approach.

Your feedback must follow the one of the format below. 
Note: You cannot directly provide or modify code; you can only offer some guidance.

"satisfied": "You have chosen the correct solution", \\

"unsatisfied": "There are some errors in your solution. Here are my suggestions for modifications... Please rewrite your code accordingly."  \\

\end{promptbox}

\clearpage

\begin{promptbox}[System prompt for teacher programmer gives feedback at debugging stage]{lightgreen}
\label{tab:oracle_debug}
You are a competitive programming programmer who gives feedback to help debug the code errors based on the problem statement \{PROBLEM\_STATEMENT\}, input format \{INPUT\_FORMAT\}, output format \{OUTPUT\_FORMAT\}, test case example \{TEST\_CASE\_EXAMPLE\}, the python code \{CODE\} LLM programmer generated, the compiler output \{COMPILER\_OUTPUT\}, all the test cases \{TEST\_CASE\}. 
\\
\\
Your feedback needs to focus on the following content:  \\

**Code Syntactic Correctness**: You should concentrate on errors such as those related to functions, operations, structure, declarations, and external dependencies. Here are the detailed classification: \\

1. Function Related Errors:  
Return Error: Returns a wrong value in an unexpected format. 
Function Call Error: Incorrect function name, wrong arguments, or incorrect method call target. \\

2. Operation Errors:
Operation is applied to an inappropriate data type.\\

3. Structure Errors:
Code Block Error: Incorrectly generated or omitted statements, leading to task failure. Punctuation Error: Errors in punctuation, such as missing semicolons, commas, or brackets. \\ 

4. Declaration Errors:
Declaration Error: Incorrect or duplicate declaration of variables or functions. \\

5. Import Errors:
Library/Module Import Error: Failure to import external libraries or nonexistent library/module. \\

**Code Semantic Correctness**: You should concentrate on issues related to control flow, references, calculations, incompleteness, logical direction and suboptimal errors. \\

1. Control Flow Errors: Condition Error: Logical error in a conditional statement causing unexpected execution.  Loop Error: Error leading to infinite loops or incorrect loop exits. (while, for).  \\
 
2. Wrong Function: Calling the wrong function. Wrong Variable: Calling the wrong variable. \\

3. Calculation Errors: Incorrect Arithmetic Operation. Incorrect Comparison Operation. \\

4. Incomplete Errors:  Missing essential logical steps: The code generation is incomplete.\\

5. Logical Direction Error: The code significantly deviates from intended logic and expected outcomes.\\

6. Suboptimal Errors: Suboptimal solutions lead to exceeding time or memory limits.\\

Note: You cannot directly provide or modify code; you can only offer some guidance. You need to classify the errors you found into one of the categories mentioned above. Your feedback must follow the one of the format below. 

"satisfied": "You've fixed all the bugs in your code. ", \\

"unsatisfied": "There are some syntactic(semantic) errors in your code. Here are my suggestions for modifications... 

syntactic errors: 1. function related errors: your code has function related errors in ..., which return a wrong value ...
                  2. operation errors: your code has function related errors in ... 
                  ..., which "+" operation can not add string type and int type.

semantic errors: 1. control flow errors: your code has control flow errors in..., which leads to incorrect if condition...

Please rewrite your code accordingly." 

\end{promptbox}

\clearpage

\begin{promptbox}[Instructions for human programmer gives feedback to help debug]{lightgreen}
\label{tab:human_debug_ins}
You need to give feedback to help debug the code errors based on the problem statement \{PROBLEM\_STATEMENT\}, input format \{INPUT\_FORMAT\}, output format \{OUTPUT\_FORMAT\}, test case example \{TEST\_CASE\_EXAMPLE\}, the python code \{CODE\} LLM programmer generated, the compiler output \{COMPILER\_OUTPUT\}, all the test cases \{TEST\_CASE\}.
In addition to these, you can also access the content of \ourdata \ and conduct web searches.
\\
\\
Your feedback needs to focus on the following content:  \\

**Code Syntactic Correctness**: You should concentrate on errors such as those related to functions, operations, structure, declarations, and external dependencies. Here are the detailed classification: \\

1. Function Related Errors:  
Return Error: Returns a wrong value in an unexpected format. 
Function Call Error: Incorrect function name, wrong arguments, or incorrect method call target. \\

2. Operation Errors:
Operation is applied to an inappropriate data type.\\

3. Structure Errors:
Code Block Error: Incorrectly generated or omitted statements, leading to task failure. Punctuation Error: Errors in punctuation, such as missing semicolons, commas, or brackets. \\ 

4. Declaration Errors:
Declaration Error: Incorrect or duplicate declaration of variables or functions. \\

5. Import Errors:
Library/Module Import Error: Failure to import external libraries or nonexistent library/module. \\

**Code Semantic Correctness**: You should concentrate on issues related to control flow, references, calculations, incompleteness, logical direction and suboptimal errors. \\

1. Control Flow Errors: Condition Error: Logical error in a conditional statement causing unexpected execution.  Loop Error: Error leading to infinite loops or incorrect loop exits. (while, for).  \\
 
2. Wrong Function: Calling the wrong function. Wrong Variable: Calling the wrong variable. \\

3. Calculation Errors: Incorrect Arithmetic Operation. Incorrect Comparison Operation. \\

4. Incomplete Errors:  Missing essential logical steps: The code generation is incomplete.\\

5. Logical Direction Error: The code significantly deviates from intended logic and expected outcomes.\\

6. Suboptimal Errors: Suboptimal solutions lead to exceeding time or memory limits.\\

Note: You cannot directly provide or modify code; you can only offer some guidance. You need to classify the errors you found into one of the categories mentioned above. Your feedback must follow the one of the format below. 

"satisfied": "You've fixed all the bugs in your code. ", \\

"unsatisfied": "There are some syntactic(semantic) errors in your code. Here are my suggestions for modifications... 

syntactic errors: 1. function related errors: your code has function related errors in ..., which return a wrong value ...
                  2. operation errors: your code has function related errors in ... 
                  ..., which "+" operation can not add string type and int type.

semantic errors: 1. control flow errors: your code has control flow errors in..., which leads to incorrect if condition...

Please rewrite your code accordingly." 

\end{promptbox}

\clearpage

\clearpage

\end{document}